\documentclass{article} 
\usepackage{iclr2026_conference,times}
\usepackage{graphicx}
\usepackage{multirow}
\usepackage{amssymb}
\usepackage{booktabs}
\usepackage{multirow}
\usepackage{makecell}
\usepackage{array}
\usepackage{adjustbox}  
\usepackage[table]{xcolor}
\arrayrulecolor{black} 
\usepackage{algorithm}      
\usepackage{algorithmic} 
\usepackage{float}

\usepackage{amsmath,amsfonts,bm}

\def\eqref#1{equation~\ref{#1}}

\def\1{\bm{1}}

\DeclareMathAlphabet{\mathsfit}{\encodingdefault}{\sfdefault}{m}{sl}
\SetMathAlphabet{\mathsfit}{bold}{\encodingdefault}{\sfdefault}{bx}{n}

\usepackage{hyperref}
\usepackage{url}
\usepackage{booktabs}
\usepackage[margin=2.2cm]{geometry}
\usepackage{graphicx}
\usepackage{booktabs} 
\usepackage{caption} 
\usepackage{pifont}
\usepackage{listings}
\usepackage{xcolor}
\usepackage{longtable}
\usepackage{array}
\usepackage{listings}
\usepackage{color}
\definecolor{codegray}{rgb}{0.95,0.95,0.95}
\usepackage{subcaption}

\usepackage{xcolor}
\usepackage[most]{tcolorbox}
\usepackage{listings}
\usepackage{marvosym}

\usepackage{bbding}
\usepackage[most]{tcolorbox}
\definecolor{codegray}{rgb}{0.95,0.95,0.95}
\definecolor{codegreen}{rgb}{0,0.6,0}
\definecolor{codepurple}{rgb}{0.5,0,0.5}

\lstdefinestyle{jsonstyle}{
    basicstyle=\ttfamily\small,
    backgroundcolor=\color{codegray},
    commentstyle=\color{codegreen},
    keywordstyle=\color{blue},
    stringstyle=\color{codepurple},
    numbers=none,
    frame=single,
    framexleftmargin=0pt,
    xleftmargin=0pt,
    xrightmargin=0pt,
    framerule=0pt,
    breaklines=true,
    showstringspaces=false
}

\title{LaTo:  Landmark-tokenized Diffusion Transformer for Fine-grained Human Face Editing}

\iclrfinalcopy

\author{Zhenghao Zhang\textsuperscript{1}\thanks{These authors contributed equally to this work.}  \qquad Ziying Zhang\textsuperscript{2}$^*$ \qquad Junchao Liao\textsuperscript{1}$^*$ \qquad  Xiangyu Meng\textsuperscript{1} \qquad Qiang Hu\textsuperscript{2}  
\\ \textbf{Siyu Zhu\textsuperscript{3} \qquad Xiaoyun Zhang\textsuperscript{2}\textsuperscript{\Letter} \qquad Long Qin\textsuperscript{1}\textsuperscript{\Letter} \qquad Weizhi Wang\textsuperscript{1}}
\\ \textsuperscript{1} Alibaba Cloud Computing \textsuperscript{2} Shanghai Jiao Tong University \textsuperscript{3} Fudan University}

\begin{document}

\maketitle

\begin{abstract}

Recent multimodal models for instruction-based face editing enable semantic manipulation but still struggle with precise attribute control and identity preservation. Structural facial representations such as landmarks are effective for intermediate supervision, yet most existing methods treat them as rigid geometric constraints, which can degrade identity when conditional landmarks deviate significantly from the source (e.g., large expression or pose changes, inaccurate landmark estimates). To address these limitations, we propose LaTo, a landmark-tokenized diffusion transformer for fine-grained, identity-preserving face editing. Our key innovations include: (1) a landmark tokenizer that directly quantizes raw landmark coordinates into discrete facial tokens, obviating the need for dense pixel-wise correspondence; (2) a location-mapped positional encoding and a landmark-aware classifier-free guidance that jointly facilitate flexible yet decoupled interactions among instruction, geometry, and appearance, enabling strong identity preservation; and (3) a landmark predictor that leverages vision–language models to infer target landmarks from instructions and source images, whose structured chain-of-thought improves estimation accuracy and interactive control. To mitigate data scarcity, we curate HFL-150K, to our knowledge the largest benchmark for this task, containing over 150K real face pairs with fine-grained instructions. Extensive experiments show that LaTo outperforms state-of-the-art methods by 7.8\% in identity preservation and 4.6\% in semantic consistency. Code and dataset will be made publicly available upon acceptance.

\end{abstract}

\section{Introduction}


Generating photorealistic facial images~\citep{lin2025ai} with controllable expressions, head pose, and other attributes while preserving subject identity remains a core challenge in face editing~\citep{preechakul2022diffusion, zhang2025museface}. These capabilities are critical for applications such as virtual avatar creation, digital human synthesis, and identity-preserving facial modifications. Recent proprietary multimodal models (e.g., SeedEdit3~\citep{wang2025seededit30fasthighquality}, Step1X-Edit~\citep{liu2025step1x}, FLUX.1-Kontext~\citep{labs2025flux}) have markedly advanced instruction-based image editing. Leveraging large-scale vision-language modeling~\citep{li2024hunyuanditpowerfulmultiresolutiondiffusion, deng2025emerging}, they deliver higher-fidelity edits across diverse scenarios than prior face editing methods~\citep{liu2022towards,pernuvs2023maskfacegan,cheng20243d}. Predictably, to enable fine-grained control, users typically must provide detailed, standardized textual descriptions (e.g., “turn the subject’s head 45° to the left and make the facial expression slightly happy”). However, existing models~\citep{liu2025step1x, xiao2025omnigen, labs2025flux} exhibit limitations in accurate instruction following and identity preservation during in-context generation. We attribute these inconsistencies to their exclusive reliance on high-level semantic encoders, which struggle to capture the structural facial cues required for precise control.

\begin{figure}[t!]
    \centering
    \includegraphics[width=0.95\linewidth]{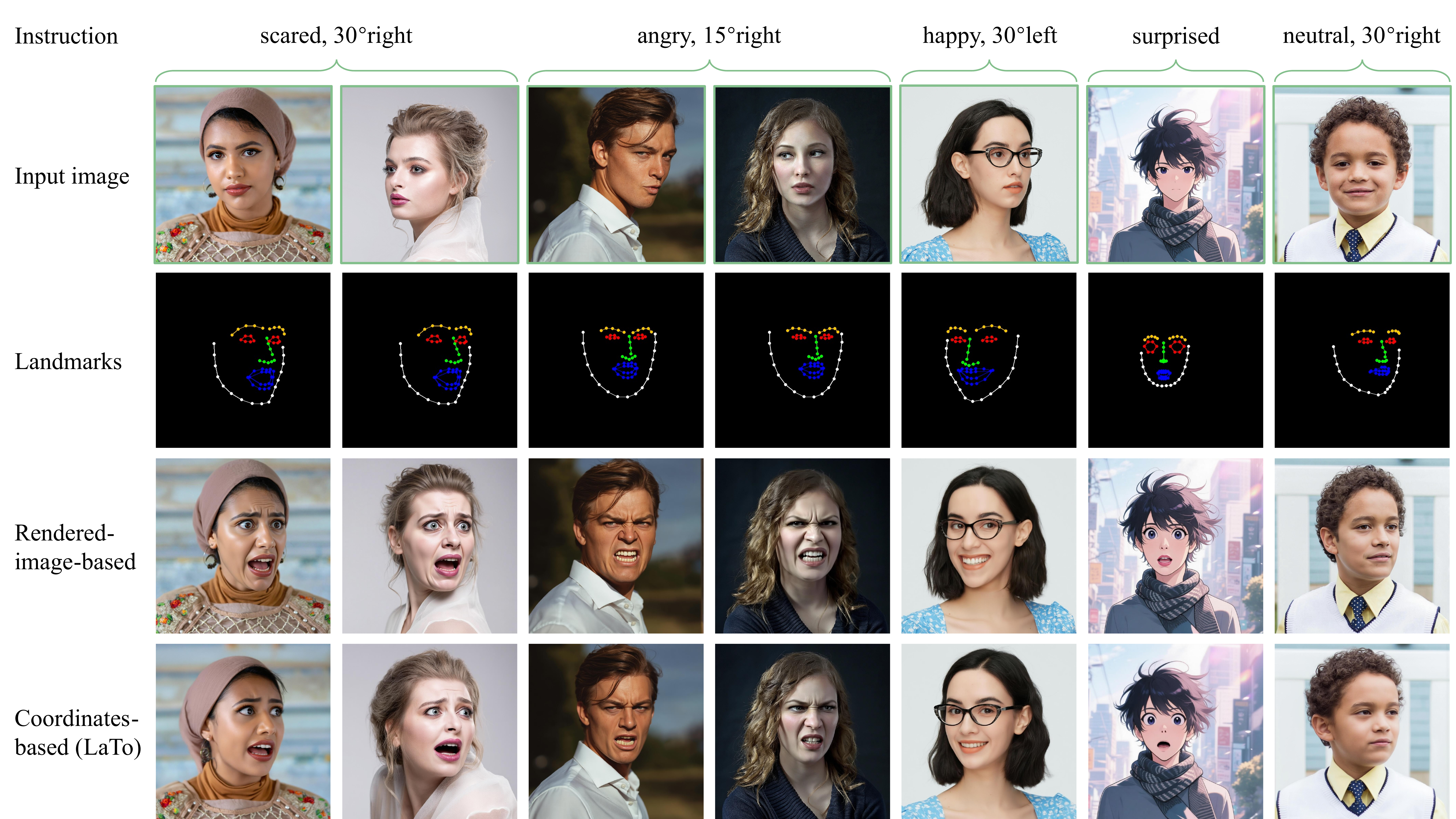}
    \caption{Landmark tokenization in LaTo preserves identity and produces natural results, whereas pixelwise alignment baselines rigidly follow the rendered landmark image and often lose identity under cross-identity landmark conditions (first four columns) or when self-identity landmarks differ substantially from the source.}
    \label{fig:first_pic}
\end{figure}

A common strategy for improving edit fidelity is to employ facial landmarks as an intermediate structural prior~\citep{yang2020generative,wei2025towards,liang2024generalizable}. Unlike text prompts, landmarks~\citep{li2022towards,sun2024lafs} impose explicit geometric constraints via precise 2D coordinates of key facial features (eyes, nose, mouth), thereby localizing edits to the appropriate regions.
However, most existing approaches are built on GANs~\citep{goodfellow2020generative} or UNet-based~\citep{ronneberger2015u} diffusion models and transfer poorly to modern Diffusion Transformer (DiT)~\citep{peebles2023scalable} due to fundamental architectural differences. Recent DiT-based editors like OminiControl~\citep{tan2024ominicontrol} and OmniGen~\citep{xiao2025omnigen} adopt a general-purpose control strategy for face editing: they rasterize landmarks into 2D images, encode them via Variational Autoencoder~(VAE)~\citep{kingma2022autoencodingvariationalbayes} to obtain dense visual tokens, and use these as in-context guidance. Despite improving identity preservation, this strategy introduces two core limitations: (1) conditioning on rendered landmark images encourages pixel-wise copying of fixed facial shapes rather than geometric reasoning, leading to identity drift and artifacts when the conditional landmarks substantially deviate from the source in shape or position, as shown in Figure~\ref{fig:first_pic}; and (2) because self-attention scales quadratically with sequence length~\citep{huang2024smartedit,avrahami2025stable}, appending long dense visual tokens to diffusion tokens incurs prohibitive memory and compute costs, limiting practical applicability in complex scenarios.

In this paper, we present LaTo, a landmark-tokenized Diffusion Transformer for complex facial editing. Instead of relying on dense pixelwise landmark renderings, we introduce a landmark tokenizer that directly quantizes landmark coordinates into discrete facial tokens. The tokenizer adopts a VQVAE–style codebook~\citep{van2017neural}, mapping coordinate inputs to embeddings with the same dimensionality as image tokens, faithfully preserving facial structure. To route sparse, spatially discontinuous landmark tokens to their target facial regions, we design a location-mapping positional encoding that anchors each token to its physical location in the latent grid, ensuring precise regional guidance in the generated image. Following Step1X-Edit, we integrate these sparse landmark tokens as contextual inputs for unified token processing within DiT blocks, enabling flexible yet decoupled interactions among geometry, appearance, and instruction while maintaining high efficiency, strong identity preservation, and semantic consistency. We further introduce landmark-aware classifier-free guidance to balance visual quality and geometric fidelity.

\begin{figure}[!t]
    \centering
    \includegraphics[width=0.9\linewidth]{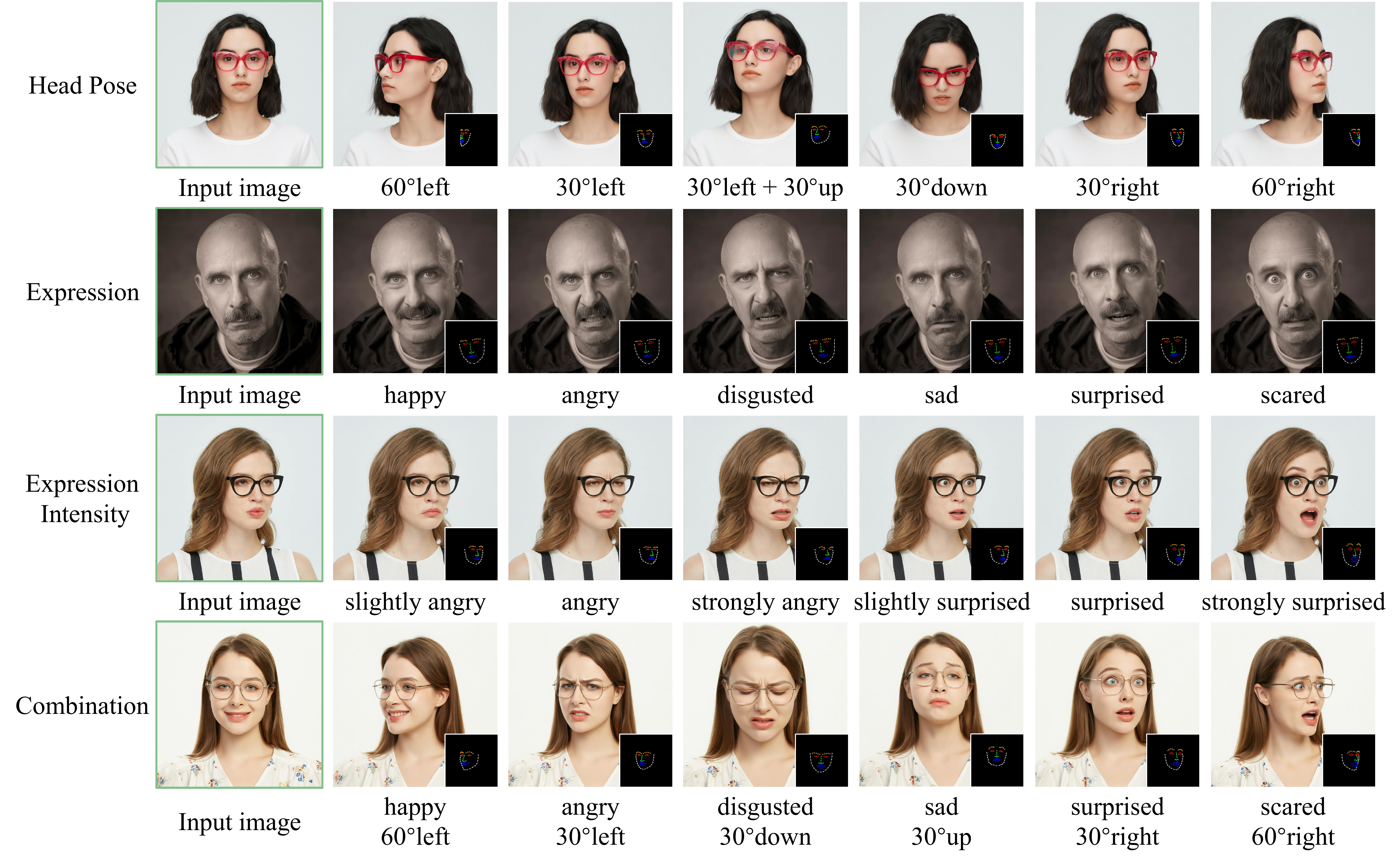}
    \caption{LaTo enables fine-grained facial expression editing, parametric head-pose editing, or their combination. The small images visualize generated landmarks via landmark predictor, enabling intuitive control signal acquisition.}
    \label{fig:show}
\end{figure}

To address training data limitations, we develop an automated synthesis-and-curation pipeline to construct HFL-150K, a large-scale dataset of more than 150,000 face editing pairs with diverse attributes and strict identity consistency. Each pair is annotated with fine-grained editing instructions and high-precision facial landmarks, providing the rich supervision necessary to fully realize LaTo’s capabilities. At inference, supplying precise landmark inputs can be impractical for end users. To alleviate this requirement, we introduce a landmark predictor that employs a vision–language model (VLM) to infer target landmarks from the source image and textual instruction using a structured chain‑of‑thought. We collect a set of high-quality instruction–landmark annotations with explicit change magnitudes and fine-tune a lightweight VLM, substantially improving landmark estimation accuracy and usability. Equipped with HFL-150K and the landmark predictor, LaTo achieves state-of-the-art face editing performance, particularly in semantic consistency and identity preservation. Our contributions can be summarized as follows:
\begin{itemize}
\item We introduce HFL-150K, a face-editing dataset comprising over 150K face pairs annotated with fine-grained editing instructions. To the best of our knowledge, it is the largest resource in this area.

\item We present LaTo, the first landmark-tokenized diffusion transformer for precise face editing that integrates (i) landmark tokenization of raw coordinates, (ii) location-mapping positional encoding, and (iii) landmark-aware classifier-free guidance. This design affords flexible geometric control, strong identity preservation, and reduced computational cost compared with rendered-image conditioning.

\item We develop a landmark predictor—a lightweight VLM that infers target landmarks from a source image and an editing instruction, bridging semantics and precise facial geometry for intuitive user control.

\item As shown in Figure~\ref{fig:show}, leveraging HFL-150K and the Landmark Predictor, LaTo delivers precise facial attribute control under complex, fine-grained instructions and achieves state-of-the-art performance.

\end{itemize}

\section{Related Work}
\subsection{Instruction-based Image Editing Models}
Diffusion models have become the defacto paradigm for high fidelity text to image synthesis and underpin many instruction driven editing systems. Existing approaches fall into two groups. Training-free methods manipulate the reverse process through latent inversion~\citep{tumanyan2023plug,rombach2022high,mokady2023null} or attention control~\citep{cao2023masactrl,wang2024taming}. These methods are efficient but often fail on complex or spatially constrained edits. Training-based methods fine tune on large scale paired image data and achieve stronger results. InstructPix2Pix~\citep{brooks2023instructpix2pix} pioneered synthetic supervision, while MGIE~\citep{fu2023guiding} and Emu Edit~\citep{sheynin2024emu} incorporate VLM to improve instruction grounding. To further narrow the gap between instructions and edits, recent work couples VLM with diffusion models, including SmartEdit~\citep{huang2024smartedit}, AnyEdit~\citep{yu2025anyedit}, UltraEdit~\citep{zhao2024ultraedit}, and unified frameworks such as OmniGen~\citep{xiao2025omnigen}, BAGEL~\citep{deng2025emerging}, and ACE~\citep{han2024ace}. Another line fuses VLMs latents into diffusion decoders (DreamEngine~\citep{chen2025multimodal}, MetaQueries~\citep{pan2025transfer}, Step1X-Edit~\citep{liu2025step1x}). Generalist systems, for example GPT-4o~\citep{hurst2024gpt} and Gemini~\citep{comanici2025gemini}, also show strong vision–language coherence. Despite these advances, precise spatial alignment and identity preservation for human face editing remain challenging because most systems rely on high level semantic signals rather than explicit geometric constraints.

\subsection{Face Editing Models}
Face editing seeks to modify facial attributes while preserving identity~\citep{preechakul2022diffusion}. Recent text driven approaches, including StyleCLIP~\citep{patashnik2021styleclip} and ChatFace~\citep{yue2023chatface}, have demonstrated strong qualitative performance.  Nevertheless, they often produce entangled edits and unintended changes to identity or appearance, particularly under large instruction variations. To improve fine-grained control and anatomical consistency, subsequent work introduces structured geometric conditions such as face masks~\citep{zhang2025museface}, semantic layouts~\citep{mofayezi2024m}, or landmark images~\citep{li2022towards,sun2024lafs}. In advanced DiT-based general editing systems~\citep{tan2024ominicontrol,pan2025transfer}, landmarks are typically rasterized into 2D images and encoded by a visual VAE to condition the diffusion process, which strengthens geometric alignment. However, pixel-wise conditioning can encourage template copying and leads to identity drift when the target geometry differs substantially from the source. Moreover, full resolution conditionals expand the token sequence and impose high memory and computation. These limitations motivate LaTo, which directly models the relationship between landmark coordinates and target facial regions, decouples geometric structure from pixel-level appearance control.

\begin{table}[t]
\centering
\small
\definecolor{lightgray}{gray}{0.9}
\caption{Key attributes of human face editing benchmarks. HFL-150K surpasses existing face benchmarks in both scale and diversity, with unique strengths in fine-grained instruction alignment.}
\begin{tabular}{ccccccc}
\toprule
Benchmarks  & Size   & Real Image & Training & Fine-grained Instruction & Expression & Head pose \\ 
\midrule
ICE-Bench~\citep{pan2025icebenchunifiedcomprehensivebenchmark}   &   206     &   \ding{51}                      &      \ding{55}     &     \ding{55}                      &       \ding{51}     &   \ding{55}         \\
SeqDeepFake~\citep{shao2022detecting} & 49,920  &     \ding{55}                      &   \ding{51}       &       \ding{55}                    &   \ding{51}         &            \ding{55}\\
SEED~\citep{zhu2025seed}        & 91,526  &     \ding{55}       &    \ding{51}                    &        \ding{55}                   &       \ding{51}     &   \ding{51}        \\ 
\midrule
\rowcolor{lightgray}
\textbf{HFL-150K}    & 302,014 &   \ding{51}                       &   \ding{51}       &    \ding{51}                      &  \ding{51}          &    \ding{51}       \\ \bottomrule
\end{tabular}
\label{tab:Comparisons_datasets}
\end{table}

\section{Methodology}
\subsection{HFL-150K Dataset Construction}
Large-scale editing datasets have been proven critical for developing advanced editing models. In face editing, existing datasets such as SeqDeepFake~\citep{shao2022detecting} and SEED ~\citep{zhu2025seed} suffer from two fundamental limitations: (1) they rely on coarse-grained facial attribute instructions and outdated synthesis models~\citep{karras2019style, tsaban2023ledits}, resulting in unrealistic editing artifacts and limited result diversity; (2) their scale is constrained by poor-quality samples that require extensive filtering to remove invalid examples.  To address the limitations of existing benchmarks, we introduce HFL-150K, a large-scale human face editing dataset comprising 150k image-edit instruction triplets~(source, instruction, edited). As summarized in Table~\ref{tab:Comparisons_datasets}, HFL-150K is constructed through a hybrid approach combining real-world image curation and synthetic generation using advanced editing models. 

\begin{figure}[t!]
    \centering
    \includegraphics[width=1\linewidth]{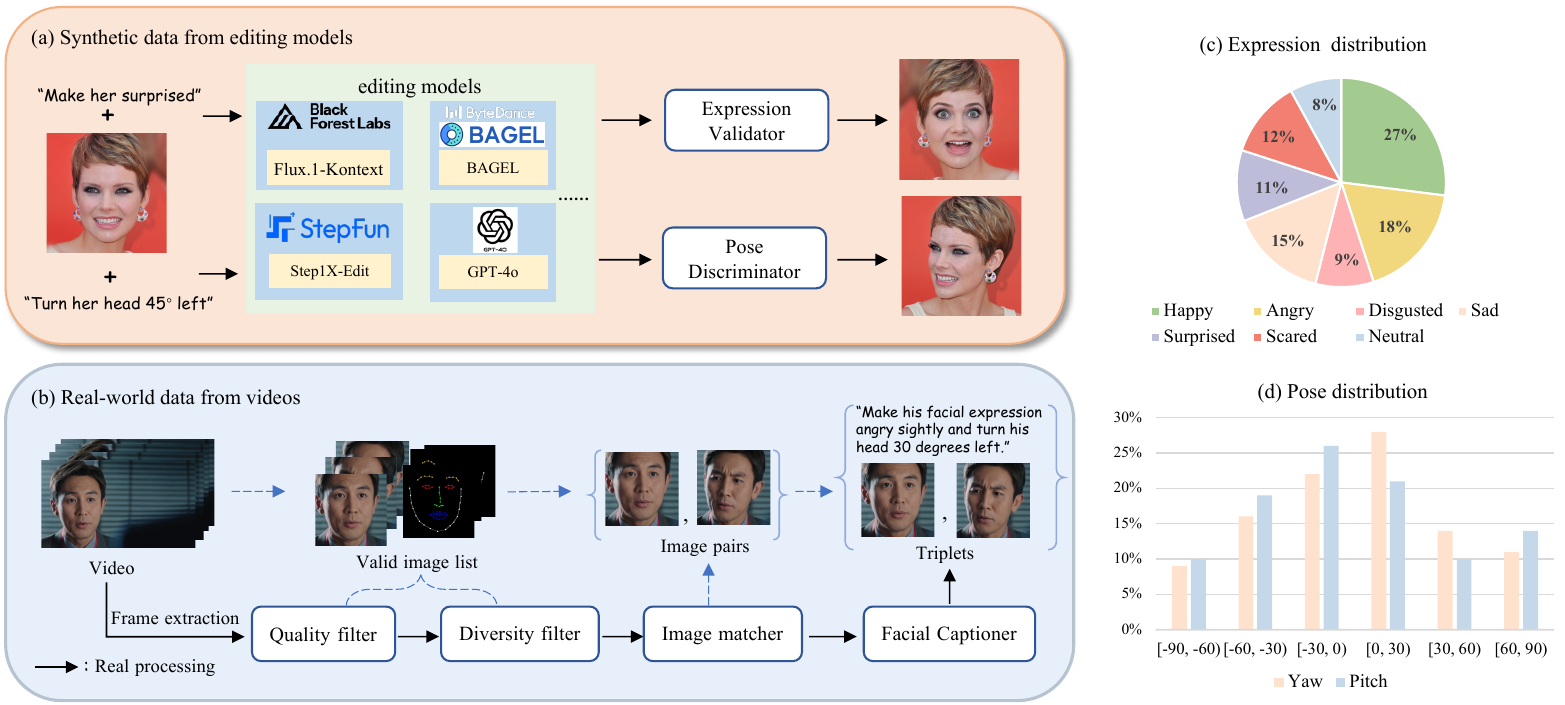}
    \caption{Data collection pipeline and statistics of HFL-150K. (a) Synthetic data generation via advanced editing models. (b) Real image pair extraction from video datasets. (c) Expression  distribution across 7 categories. (d) Head pose angles aligned with 30° motion budgets.}
    \label{fig:data_pipeline}
\end{figure}

\textbf{Fine-grained attribute definition.} We focus on two core facial editing tasks: expression editing (e.g., ``make him smile gently") and parametric head pose editing (e.g., ``rotate her head 30° left"). For expression categorization, we adopt standard emotion recognition protocols~\citep{huang2023study} to define seven canonical expressions and assign intensity levels ({slightly, normally, strongly}) based on visual saliency. For head pose parameters, we formulate spatial transformations using yaw and pitch angles with 30° as the base unit of motion amplitude, as shown in Figure ~\ref{fig:data_pipeline} (c–d).

\textbf{Synthetic data collection.} As shown in Figure~\ref{fig:data_pipeline} (a), we employ advanced editing models (Step1X-Edit, GPT-4o~\citep{hurst2024gpt}, BAGEL~\citep{deng2025emerging}, and FLUX.1-Kontext) to generate a synthetic dataset focusing on either expression or head pose edits, aligning with their single-turn training objectives. To ensure generation quality, we implement instruction-specific filtering: (1) For expression edits, an expression validator computes semantic similarity between generated outputs and input instructions using Qwen2.5-VL~\citep{bai2025qwen25vltechnicalreport}. (2) For pose edits, a pose discriminator estimates head orientation via Euler angle regression~\citep{yang2019fsa} from facial landmarks and verifies alignment with target rotations~$\theta_t$, which can be described as:
\begin{equation}
\Delta\theta = \left\| \hat{\theta} - \theta_t \right\|_2 = \sqrt{(\hat{\theta}_p - \theta_{t,p})^2 + (\hat{\theta}_y - \theta_{t,y})^2 }, 
\end{equation}
where $\hat{\theta} = (\hat{\theta}_p, \hat{\theta}_y)$ represents the estimated pitch, yaw angles. Only validated samples are retained, resulting in a \textbf{34K-sample} dataset. This synthetic dataset provides a simplified prior that enhances model interpretability.

\textbf{Real-world data collection.} As illustrated in Figure~\ref{fig:data_pipeline} (b), we further construct a real-world face subset from human-centric video datasets~\citep{li2025openhumanvid}, leveraging natural dynamics to capture intra-identity variation in expression and head pose. We apply a multi-stage filtering process comprising: (1) a quality filter using a face detector~\citep{deng2019arcface} to drop occlusions and enforce centering, a Laplacian of Gaussian (LoG) to remove motion blur, and Dover~\citep{wu2023exploring} for aesthetic and technical assessment; (2) a diversity filter combining geometric analysis (2D facial landmarks~\citep{zhu2016unconstrained}) and semantic analysis (Qwen2.5-VL for high level facial changes). Pairs with scores beyond thresholds are removed, including abnormally high values indicating copy-paste artifacts and critically low values caused by variability between different people. Finally, an image matcher with CLIP~\citep{radford2021learning} performs cross frame identity verification, removing residual pairs from different people. Through this pipeline, a total of \textbf{116K} high-quality, semantically diverse image pairs are generated, reflecting natural facial variations.

For deriving fine-grained editing instructions, we leverage a facial captioner to recognize expressions and estimate intensity. However, even advanced Qwen2.5-VL-72B shows limitations in fine-grained expression recognition. To address this, we curate a high-quality dataset and fine-tune the model to improve sensitivity to more subtle magnitudes. We manually annotate 18k samples with seven predefined expression categories and intensity levels (see Appendix for guidelines). For head pose estimation, we use both horizontal and vertical optical flow angles~\citep{karaev2024cotracker3simplerbetterpoint} to quantify motion amplitude relative to our base unit.  A unified template is designed for instruction generation:

   \textit{Make his/her facial expression \{expression-type\} \{intensity-level\} and turn his/her head \{angle-degree\}.}

\subsection{LaTo}
\begin{figure}[t]
    \centering
    \includegraphics[width=0.9\linewidth]{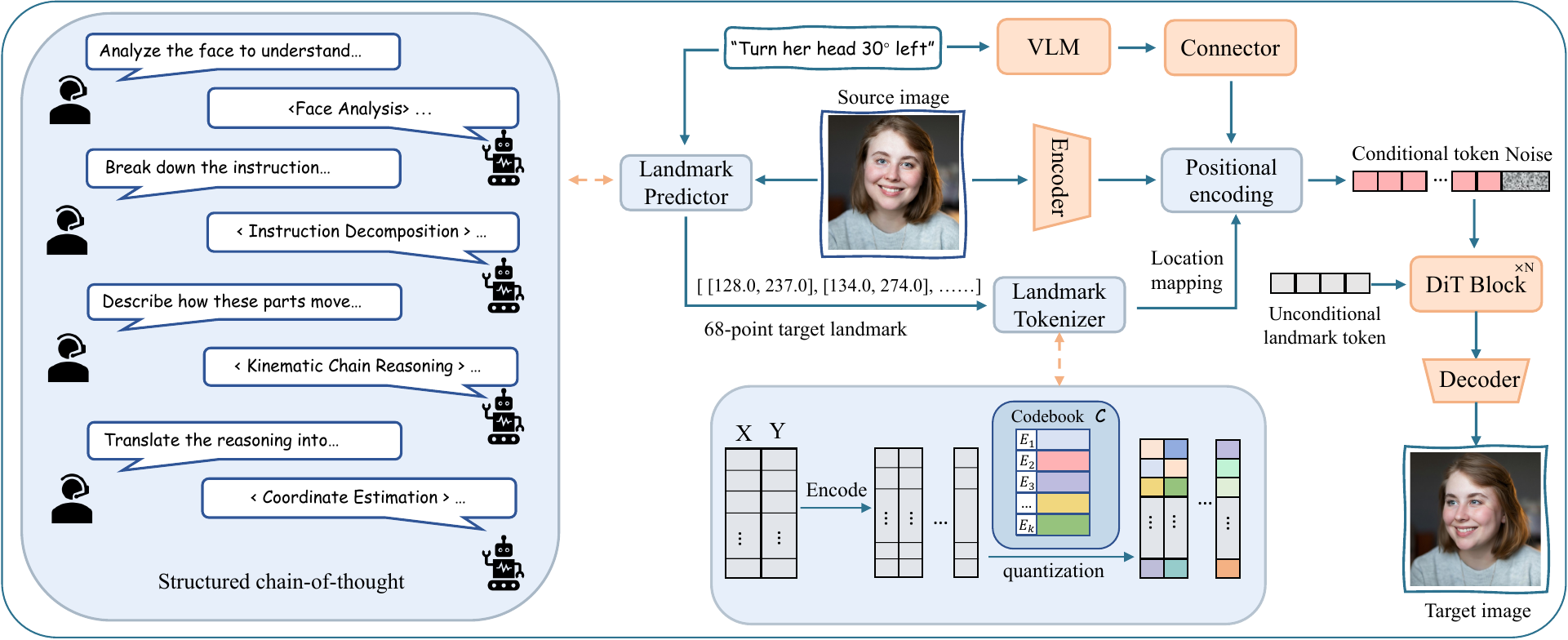}
    \caption{Overview of LaTo. The landmark predictor infers target landmarks from source image and instruction via structured chain of thought. A landmark tokenizer and visual VAE encode predicted landmarks and source image into tokens. The location-mapping positional encoding anchors each landmark token to its physical location, ensuring unified yet flexible alignment with instruction and visual tokens. The learned unconditional landmark token further guides the denoising process, keeping the edited image aligned with both the specified landmarks and instructions.}
    \label{fig:model_pipe}
\end{figure}

Building upon the Step1X-Edit, we propose LaTo, a fine-grained human face editing framework that effectively leverages compact facial tokens. As illustrated in Figure~\ref{fig:model_pipe}, LaTo achieves precise and user-friendly face editing through three core mechanisms: landmark tokenizer, multi-modal token fuser and landmark predictor. 

\subsubsection{Landmark Tokenizer}

Building on the widely adopted VQVAE architecture for discrete tokenization, the landmark tokenizer combines an encoder-decoder framework with a lightweight quantizer. Given a  raw landmark sequence $F = {\left \{ (X_{i}, Y_{i}) \right \}}_{i=1}^{n}$  (with $ n $ 2D locations), the encoder maps it into a continuous latent space $E \in  R^{n \times d} $ via residual blocks with convolutions. A quantizer then discretizes these latents through nearest-neighbor lookup in a learnable codebook $C \in R^{m \times d}$ of size $ m $, generating compact yet expressive facial tokens in a unified geometric space. The decoder, structured to mirror the encoder, reconstructs the input sequence, ensuring spatial coherence. The complete training objective combines reconstruction loss and commitment loss:
\begin{equation}
    \mathcal{L} = \|F - \hat{F}\|_{1} + \beta \| E - \operatorname{sg} \left[ C \right] \|_{2}^{2}
\end{equation}

where sg$[\cdot]$ denotes the stop-gradient operation, and $\beta$ is the weight of the commitment loss. This loss encourages faithful reconstruction while promoting effective codebook utilization, enabling the model to learn both geometric accuracy and semantic expressiveness in facial token representations.

\subsubsection{Multi-modal token fuser}
We develop a token fuser to flexibly integrate landmark, image, and semantic tokens through three components: location-mapped landmark positional encoding, unified representation and landmark-aware classifier-free guidance.

\noindent \textbf{Location-mapping landmark positional encoding.} Step1X-Edit employs 3D Rotary Positional Encoding (RoPE) ~\citep{su2024roformer} to encode spatial information for both image and text tokens.  For each position $i$ in image tokens, the 3D RoPE is computed as: 
\begin{equation}
    P_i = \text{Concat}\left(R_{T}(0),\, R_{H}\left(\left\lfloor \frac{i}{h} \right\rfloor\right),\, R_{W}(i \% h)\right),
\label{eq:rope}
\end{equation}
where $ h $ represents the height of the latent grid. Here, each $ R(\cdot) $ implements 1D rotary embeddings applied to the text, height, and width dimensions. These embeddings are independently repeated across spatial axes to model positional relationships between tokens. To maintain spatial consistency, we design a location mapping mechanism that links landmark tokens to their physical location in the latent grid:

\begin{equation}
    P_i = \text{Concat}\left(R_{T}(0),\, R_{H}\left(y_{i}\right),\, R_{W}(x_{i})\right),
\label{eq:our_rope}
\end{equation}
where $(x_i, y_i)$ denote downsampled coordinates of landmark points from the original image. This ensures each compressed representation accurately guides its corresponding area, preserving conditioning fidelity. Our experiments~\ref{sec:ab} demonstrate that, in the absence of this correction, the model struggles to learn spatial relationships between landmark inputs and generated tokens, leading to blurry or misaligned facial features.

\noindent \textbf{Unified representation.} Following Step1X-Edit, we treat all modal tokens as a unified representation. A trainable facial landmark adapter first projects facial tokens into the same latent space as noisy  tokens, denoted as $ z_f \in \mathbb{R}^{l_f \times d} $. The input is formulated as:  
\begin{equation}
\mathbf{Z} = \text{Concat}(z_{\text{t}}, z_{\text{s}}, z_f, z_{\text{n}}) \in \mathbb{R}^{(l_t + l_s + l_f + l_n) \times d},
\end{equation}
where $ z_{\text{t}} \in \mathbb{R}^{l_t \times d}$, $ z_{\text{s}} \in \mathbb{R}^{l_s \times d}$, and $ z_{\text{n}} \in \mathbb{R}^{l_n \times d}$ denote semantic text tokens, source visual tokens, and noisy image tokens, respectively. The multi-modal attention mechanism generates query and key via: 
\begin{equation}
\begin{aligned}
\mathbf{P} &= \text{Concat}(P_{\text{t}}, P_{\text{s}}, P_f, P_{\text{n}}) \\
\mathbf{Q} &= \text{RoPE}(W_q(\mathbf{Z}), \mathbf{P}), \quad \mathbf{K} = \text{RoPE}(W_k(\mathbf{Z}), \mathbf{P}),
\end{aligned}
\end{equation}
with $ P_{\text{t}}, P_{\text{s}}, P_{\text{n}} $ derived from Equation~\ref{eq:rope} and $ P_f $ from Equation~\ref{eq:our_rope}. This formulation enables flexible token interactions via DiT’s multi-modal attention mechanism, allowing direct relationships between any token pair without rigid spatial constraints. Given the facial landmark token length $l_{f}= 68$ is significantly smaller than noisy image tokens ($l_{n}=1024$), this approach maintains computational efficiency comparable to the baseline model.

\noindent \textbf{Landmark-aware Classifier-free guidance.} To balance image quality and landmark fidelity, we introduce landmark-aware classifier-free guidance (CFG). In conventional image-conditioned pipelines, the unconditional branch is obtained by feeding a zero image. By analogy, zeroing landmark coordinates for the unconditional path, especially at high CFG weights, often makes the model copy the reference face and suppress appearance variations that should covary with landmarks.  We argue that zeroed landmark embeddings do not encode an unconstrained geometric state and they conflict with the physical dynamics of facial motion. This yields an undesired coupling between landmark conditions and the generated content even in the unconditional branch. Inspired by MTVCrafter~\citep{ding2025mtvcrafter4dmotiontokenization}, we train learnable unconditional tokens that replace the position-aware landmark tokens during unconditional training passes. This produces a semantically meaningful unconditional distribution and improves robustness.

\subsection{Landmark Predictor}

To enable intuitive interaction and improve landmark estimation, we fine-tune Qwen2.5-VL-3B into a landmark predictor (LP). Given a source image and an instruction, LP generates target landmark coordinates via a structured CoT. The CoT supervision comprises four stages: (1) initial state analysis, characterizing the starting pose, expression, and landmark alignment; (2) instruction decomposition, breaking the instruction into primary anatomical motions; (3) kinematic-chain reasoning, separating rigid motions (head rotation, translation) from non-rigid deformations (muscle-driven expression changes); and (4) coordinate estimation, mapping these components to numerical displacements on a normalized 512$\times$512 canvas and producing a canonical, machine-parsable list of $(X, Y)$ pairs. We generated CoT traces for 23,145 triplets sampled from HFL-150K using a rule-guided pipeline that ingests the source image, instruction and target landmarks, and after manual verification retained 19,398 high-quality examples for fine-tuning. During training, the visual and textual inputs are encoded and fused in the multimodal transformer and the model is optimized with next-token supervision to produce the structured CoT token sequence. Coordinates are normalized and encoded with a compact tokenization scheme and a fixed output grammar to improve numeric fidelity. At inference, smoothing and geometric sanity checks are applied to preserve identity-consistent rigid distances. This design delivers interpretable, numerically precise landmark predictions that connect robust instruction understanding to explicit geometric control for downstream face editing.  Detailed CoT procedures are provided in the Appendix.

\section{Experiments}
\subsection{Experimental Setup}
\label{sec:experiment_setup}


\begin{table}
\centering
\setlength{\tabcolsep}{3.5pt}
\renewcommand{\arraystretch}{1.0}
\small
\captionof{table}{Quantitative evaluation of state-of-the-art editing methods on HFL-150K test set and face attribute editing subsets from GEdit-Bench/ICE-Bench.
† Indicates models fine-tuned on HFL-150K training set.}
\begin{tabular}{ccccccccc}
\toprule
\multicolumn{1}{c|}{\multirow{2}{*}{Method}}                            & \multicolumn{4}{c|}{HFL-150K}    & \multicolumn{4}{c}{GEdit\&ICE-Bench(Subset)} \\ \cmidrule{2-9} 
\multicolumn{1}{c|}{}                                                   & SC~$\uparrow$ & VQ~$\uparrow$ & NA~$\uparrow$ & \multicolumn{1}{c|}{IP~$\uparrow$} & SC~$\uparrow$         & VQ~$\uparrow$         & NA~$\uparrow$         & IP~$\uparrow$         \\ \midrule
Instruct-PixPix~\citep{brooks2023instructpix2pix}   &    0.518                &    0.582                   &    0.675            &     0.381                          &    0.573                &    0.567                   &    0.643            &     0.405             \\
AnyEdit~\citep{yu2025anyedit}                     & 0.612                     & 0.641                      &    0.702            &  0.446                          &     0.669                     & 0.654                      &    0.676            &  0.479          \\
OmniGen~\citep{xiao2025omnigen}       & 0.737                     &0.688                       & 0.731               &   0.503                        &        0.755                     &0.707                       &  0.697              &   0.536                    \\
Bagel~\citep{deng2025emerging}                   & 0.786                      &  0.709                     &  0.759              &  0.539                         &      0.797                      &  0.718                     &  0.733              &  0.579          \\
Step1X-Edit~\citep{liu2025step1x}                & 0.751                     & 0.706                      &  0.732              &  0.518                        &        0.767                     & 0.694                      &  0.705              &  0.541            \\
Step1X-Edit† ~\citep{liu2025step1x}                 &  0.804                    &0.725                       &  0.801               & 0.571                         &     0.803                    &\textbf{0.732}                       &  0.788               & 0.594            \\
FLUX.1-Kontext~\citep{labs2025flux}              &  0.712                    &0.720                       &   0.779             &    0.556                          &    0.749                    &0.693                       &   0.751             &    0.576           \\
FLUX.1-Kontext† ~\citep{labs2025flux}              & 0.786                     & 0.737                      &   \textbf{0.816}             &    0.593   &                  0.801                     & 0.713                      &   0.771             &    0.609           \\
\textbf{LaTo (ours)}                                           &  \textbf{0.832}                    &   \textbf{0.749}                    &    0.805            &     \textbf{0.634}                          &   \textbf{0.829}                   &   0.724                    &    \textbf{0.793}            &     \textbf{0.651}        \\
\bottomrule
\end{tabular}
\label{tab:compare}
\end{table}

\textbf{Implement details.} 
The landmark tokenizer is trained from scratch on HFL-150K with an 8,192-entry codebook and 3,072-dimensional codes matched to the Step1X-Edit hidden size. Training uses 8 NVIDIA A100 GPUs with batch size 128 per GPU for 100k iterations, and unused codes are reset every 50 steps to prevent saturation. For LaTo, we fine-tune the base model with LoRA (rank 64) and add an unfrozen linear landmark adapter. The model is trained on 16 NVIDIA A100 GPUs with total batch size 32, learning rate $1 \times 10^{-4} $. We replace conditional landmark tokens with unconditional ones with probability 0.1 to encourage diversity, and train for a total of 40k iterations.

\textbf{Dataset.} We split HFL-150K into 150,000 training samples and 1,007 test samples. For the test set, stratified sampling ensures diverse coverage of expression types, head poses, and their combinations. We additionally perform manual validation to ensure all samples include high-quality instructions and image–landmark pairs. We also curate an auxiliary test set by selecting face-attribute editing samples from GEdit-Bench~\citep{liu2025step1x} and ICE-Bench~\citep{pan2025icebenchunifiedcomprehensivebenchmark}, yielding 103 samples for further evaluation. Detailed curation protocols are provided in the Appendix.

\textbf{Metrics.} 
We propose a four-criteria framework to evaluate both identity preservation and accuracy of requested modifications, comprising Semantic Consistency (SC), Visual Quality (VQ), Natural Appearance (NA), and Identity Preservation (IP). SC, VQ, and NA are scored by Qwen2.5-VL-72B using a normalized $[0, 1]$  scale via visual reasoning. For IP, we first use ArcFace similarity $s_{arc}$~\citep{deng2019arcface}, but some methods inflate it by copying or barely altering the source. We therefore define a rectified score that penalizes such cases: Qwen2.5-VL parses the source face and predicts expected edit amplitude $\varphi_{ins}$from the instruction, while the actual amplitude $\varphi_{real}$ is derived from SSIM between source and edit. The rectified IP score calculated as:

\begin{equation}
s_{rip} = \max(0, s_{arc} -  (\frac{\varphi_{ins} - \varphi_{real} }{\varphi_{ins} + \epsilon } )^{2}) ,
\end{equation}
we provide the pseudocode in the Appendix and confirm that this metric achieves better alignment with human preference through a user study.

\subsection{Qualitative and Quantitative Analysis} 

\begin{figure}[!t]
    \centering
    \includegraphics[width=1\linewidth]{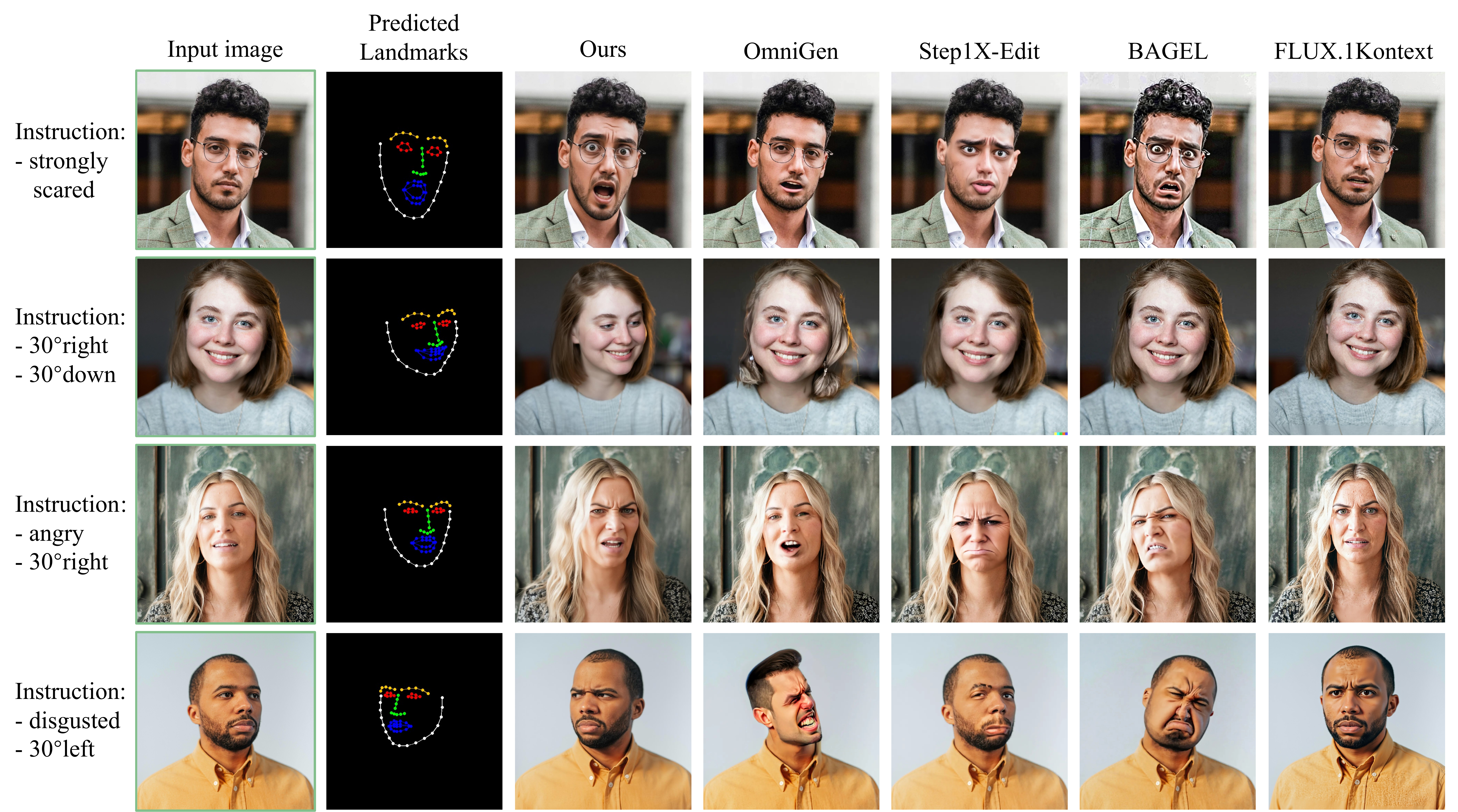}
    \caption{Qualitative comparison with state-of-the-art image editing methods. 
    }
    \label{fig:comparison_main}
\end{figure}

We benchmark LaTo against state-of-the-art methods on two datasets: HFL-150K (fine-grained edit instructions) and the face-attribute splits of GEdit-Bench/ICE-Bench (global descriptions). For fairness, we fine-tune Step1X-Edit and Kontext on HFL-150K using their official implementations. Table~\ref{tab:compare} reports statistically significant gains across all metrics. We observe 5.3\% and 7.4\% relative improvements in SC on HFL-150K for the two approaches, suggesting that our dataset better reflects real-world diversity for this task. On HFL-150K, LaTo surpasses the second-best method, Bagel, by 4.6\% SC, and LaTo-IP exceeds FLUX.1-Kontext by 7.8\%. On GEdit-Bench/ICE-Bench, LaTo outperforms Bagel by 7.2\%, demonstrating stronger identity preservation. Despite sharing the same training data and base models as Step1X-Edit†, LaTo achieves a 2.9\% average absolute improvement across all metrics, validating the effectiveness of our landmark tokenization design. Qualitative results (Figure~\ref{fig:comparison_main}) show superior pose accuracy and identity consistency, while maintaining photorealism under large expression changes where most baselines introduce cartoon-like or synthetic artifacts.

\subsection{Ablation study}
\label{sec:ab}

\begin{table}[!t]
\centering
\setlength{\tabcolsep}{2.5pt}
\renewcommand{\arraystretch}{1.05}
\small
\caption{Ablation study on landmark condition types and landmark positional encoding. We calculate the L1 distance between the edited image and provided landmark as the Landmark Error to evaluate the landmark alignment precision.}
\begin{tabular}{ccccccccc}
\toprule
Landmark type                                                                             & Encoder                                                                        & Additional Setting & SC~$\uparrow$ & VQ~$\uparrow$ & NA~$\uparrow$ & IP~$\uparrow$ & Latency(s)~$\downarrow$ & Landmark Error~$\downarrow$ \\ 
\midrule
/ & / &  / & 0.804                    &0.725                       &  0.801               & 0.571                         &     \textbf{49.6}  & - \\
\hline
\multirow{2}{*}{\begin{tabular}[c]{@{}c@{}}rendered image\end{tabular}} & \multirow{2}{*}{visual VAE}                                                    & /                 &  0.821  &  0.712  &  0.744  &  0.584  &    83.6   &    \textbf{1.76}   \\
                                                                                           &                                                                                & compression        & 0.816   & 0.704   & 0.709   &  0.569  &    61.3 & 3.07     \\ \hline
\multirow{4}{*}{coordinates}                                                             & \multirow{4}{*}{\begin{tabular}[c]{@{}c@{}}landmark\\ tokenizer\end{tabular}} & (1)~w/o PE   &   0.654       &  0.611  &  0.630  &  0.512  &  50.7 &  58.9       \\
                                                                                           &                                                                                & (2)~RoPE         & 0.778   &  \textbf{0.751}  & 0.786   & 0.621   &   52.1  & 25.4      \\
                                                                                           &                                                                                & (3)~learnable RoPE       & 0.803    &    0.737 &  0.791  &  0.617  &  52.9 & 9.63        \\
                                                                                           &                                                                                & (4)~ours   &  \textbf{0.832}                    &   0.749                    &    \textbf{0.805}            &     \textbf{0.634}  &52.1      & 2.34   \\ 
\bottomrule
\end{tabular}
\label{tab:ab1}
\end{table}

\begin{figure}[!t]
    \centering
    \includegraphics[width=0.94\linewidth]{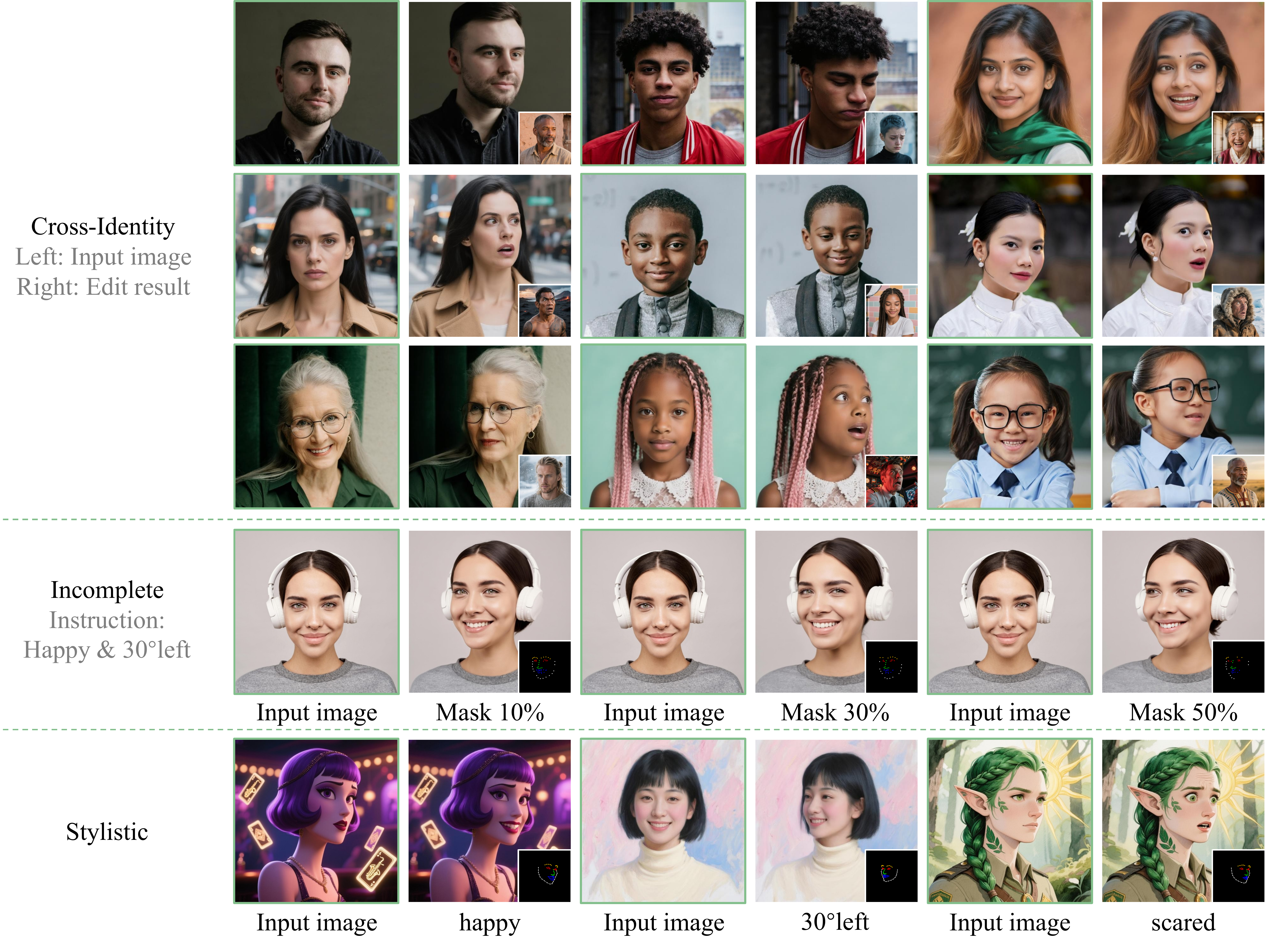}
    \caption{Qualitative results on challenging inputs, including cross-identity landmarks, incomplete landmarks, and stylistic inputs. For cross-identity cases, the corresponding driving images are shown in the bottom-right corner.}
    \label{fig:cross-id}
\end{figure}

\begin{table}[t]
\centering
\begin{minipage}[t]{0.48\linewidth}
\vspace{0pt}%
\centering
\setlength{\tabcolsep}{3.5pt}
\renewcommand{\arraystretch}{1.0}
\small
\captionof{table}{Comparison of landmark predictor performance using unified prompting conditions.}
\begin{tabular}{cc}
\toprule
Methods            & Accuracy \\ 
\midrule
Qwen2.5-VL-3B &   0.477 \\
Qwen2.5-VL-72B &   0.597 \\
Gemini 2.5 Pro &   0.613 \\
\textbf{Ours}  &   \textbf{0.730}  \\
\bottomrule
\end{tabular}
\label{tab:user_study_main}
\end{minipage}
\hfill
\begin{minipage}[t]{0.48\linewidth}
\vspace{0pt}%
\centering
\setlength{\tabcolsep}{3pt}
\renewcommand{\arraystretch}{1.0}
\small
\setlength{\tabcolsep}{2.5pt}
\renewcommand{\arraystretch}{1.0}
\small
\captionof{table}{Performance analysis across different CFG scales for landmark condition.}
\begin{tabular}{cccccc}
\toprule
CFG-scale           & SC~$\uparrow$ & VQ~$\uparrow$ & NA~$\uparrow$ & IP~$\uparrow$ &Landmark Error~$\downarrow$ \\ 
\midrule
1 &     0.803                   &    0.733            &     \textbf{0.816} & 0.619      & 9.63        \\
4          & \textbf{0.832}                     &       \textbf{0.749}                &   0.805             &  \textbf{0.634}        & 2.34            \\
7     &    0.821                  &  0.698                     &     0.751           &    0.616       & 1.94          \\
10   &    0.807                  &    0.656                   &   0.679             & 0.588              & \textbf{1.82}       \\ 
\bottomrule
\end{tabular}
\label{tab:cfg}
\end{minipage}
\end{table}

\textbf{Facial condition formulations.} We compare against two standard landmark conditioning schemes: (1) rendering landmarks as 2D images and extracting facial tokens with a shared VAE, and (2) downsampling landmark images with position shifting (inspired by OmininControl2~\citep{tan2025ominicontrol2efficientconditioningdiffusion}) to improve efficiency. As shown in Table~\ref{tab:ab1}, relative to the fine-tuned baseline, these image-based variants are limited: NA decreases by 5.7\%, IP increases only by 1.3\%, and compression further widens the gap despite a 26\% speedup. In contrast, our landmark tokenization models spatial relations between coordinates and facial attributes and decouples control strength, achieving 6.1\% higher NA, 1.1\% higher SC, and 5.0\% higher IP than rendered-image conditioning. It also attains a 37\% speedup, matching the baseline’s computational efficiency.

\textbf{Landmark positional encoding effectiveness.}  We investigate various positional encoding (PE) strategies for landmark tokens, including original relative encoding (RoPE), learnable RoPE, no PE, and our location-mapping RoPE. The results in Table~\ref{tab:ab1} demonstrate that: (1) No PE leads to unstable training and unnatural results due to the absence of spatial awareness; (2) Original RoPE fails to effectively capture spatial relationships between landmarks and target images, achieving a suboptimal landmark error of 25.4; (3) Learnable RoPE improves both instruction adherence and landmark conditioning but remains inferior to our approach; (4) our method provides the model with a strong geometric prior, enabling rapid geometry information extraction and achieving superior performance in both landmark conditioning and identity fidelity.

\begin{figure}[!t]
    \centering
    \includegraphics[width=0.93\linewidth]{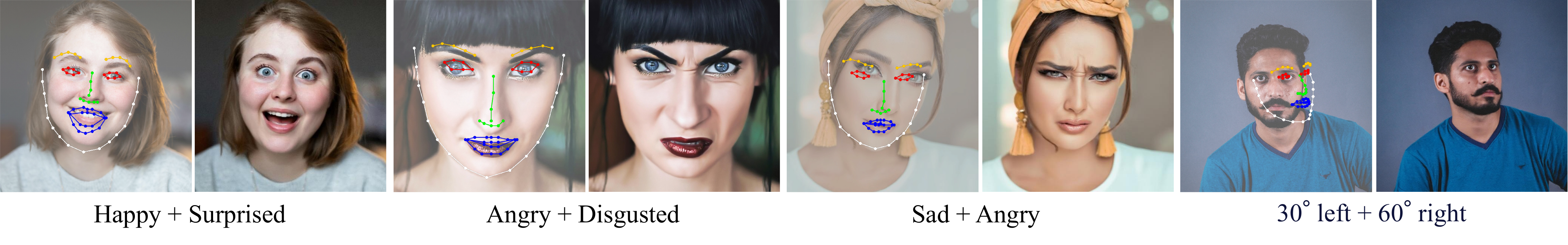}
    \caption{Visualization of the performance of landmark predictor under ambiguous instructions. Predicted landmarks are overlaid on the source images.}
    \label{fig:ambiguous_lp}
\end{figure}

\textbf{Landmark predictor accuracy evaluation.}   To evaluate the effectiveness of landmark predictor, we conducted a manual accuracy assessment against Gemini 2.5 Pro, Qwen2.5-VL-72B, and Qwen2.5-VL-3B. Specifically, we selected 50 human faces from the HFL-150K test set and randomly generated instructions for each image by altering expression, head pose angle, or their combinations (6 variations per image), resulting in 300 samples. The predicted landmarks were rendered on source images, and participants were asked to evaluate whether the predicted landmarks aligned with the given instructions, assigning a score of 0 (incorrect) or 1 (correct). We recruited 10 human evaluators and collected their results, which are presented in Table~\ref{tab:user_study_main}. Our fine-tuned model achieves the highest average accuracy among the compared models, outperforming the baseline by 25.3$\%$, demonstrating its superiority in landmark analysis. To further evaluate LP under realistic noisy instructions, we test it on composite expressions (e.g., happy + surprised) and conflicting pose instructions (e.g., turn left and turn right), as shown in Figure~\ref{fig:ambiguous_lp}. LP remains robust to instruction-level ambiguity. For composite expressions, it produces plausible intermediate states in the landmark space. For conflicting poses, the CoT module infers a coherent head orientation and often defaults to a near-frontal view when cues conflict.

\begin{figure}[!t]
    \centering
    \includegraphics[width=0.93\linewidth]{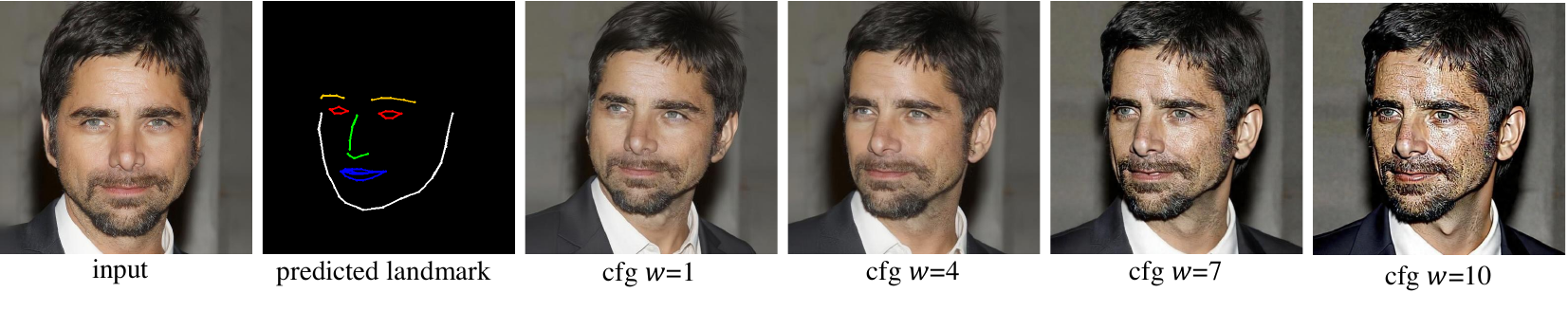}
    \caption{Visualization of the landmark-aware CFG scale $w$.}
    \label{fig:cfg}
\end{figure}

\textbf{Landmark-aware CFG scaling analysis.}  Figure~\ref{fig:cfg} and Table~\ref{tab:cfg} illustrate the impact of our CFG scale. Raising the CFG scale improves landmark alignment, yet may generate more artifacts and compromise video quality. We adopt a CFG scale of 4 as the optimal baseline configuration.

\textbf{The sensitivity to challenging inputs.} Although LaTo is primarily self-landmark-driven via the proposed LP, we also evaluate its behavior in three challenging scenarios: (1) incomplete landmarks, (2) cross-identity landmarks, and (3) stylistic inputs. As shown in Figure~\ref{fig:cross-id}, LaTo largely preserves identity under cross-identity conditioning and moderate landmark dropout, and maintains stable editing quality on non-photorealistic inputs. We attribute this to three key components. First, the landmark tokenizer disentangles explicit shape-related structure from pixel-level appearance, which stabilizes geometric reasoning and facilitates the capture of generalized expression dynamics and identity-agnostic pose variations . Second, the token fuser encourages the model to interpret driving facial geometry, editing instructions, and visual identity features in a coordinated manner, which enables driving-geometry control in an implicit, feature-level manner, rather than relying on explicit geometric following. Third, the learned unconditional landmark tokens give each landmark position a flexible way to adjust the coupling between geometry and identity, so that even when landmark absence becomes severe (about 50$\%$), LaTo, despite noticeable degradation in identity preservation, still maintains basic visual quality and key facial traits.

\section{Conclusion}
We presented LaTo, a landmark-tokenized diffusion transformer for fine-grained, identity-preserving face editing. LaTo quantizes landmark coordinates into discrete facial tokens and aligns them with image tokens via a location mapping positional encoding, which decouples geometry from appearance and enables precise control with strong identity preservation and high efficiency. A vision–language landmark predictor with structured reasoning infers target landmarks from instructions and source images, improving robustness and interactive controllability. This design removes the need for dense pixel-wise correspondence and mitigates identity drift under large pose or expression changes. To support research at scale, we curate HFL-150K, a large-scale benchmark of face pairs with fine-grained instructions, spanning real-world imagery and outputs from advanced models. Extensive experiments demonstrate that LaTo delivers state-of-the-art photorealism, semantic consistency, and computational efficiency, establishing a strong foundation for controllable, human-centric editing.

\section{Acknowledgements}
This work is supported by National Natural Science Foundation of China (62571322, 62431015, 62271308), STCSM (24ZR1432000, 24511106902, 24511106900, 22DZ2229005), 111 plan (BP0719010), and State Key Laboratory of UHD Video and Audio Production and Presentation.


\bibliography{iclr2026_conference}
\bibliographystyle{iclr2026_conference}

\appendix
\clearpage
\section*{Appendix}
\section{The Use of Large Language Models}
After completing the manuscript, we used large language models to check grammar and refine the writing.

\section{Dataset Illustration}
\subsection{More Details of filtering pipeline}
For real image pairs, the quality filter retains samples that contain a single face whose centroid lies within the central 20\% of the image, with facial landmarks covering at least 7\% of the image area, and with motion blur below a Laplacian of Gaussian (LoG) threshold of 50. Dover is applied with a threshold of 0.5 for aesthetic and technical assessment. The diversity filter then enforces both geometric and semantic coverage. For geometric diversity, we compute a composite change score combining local (eyes/eyebrows/mouth; indices 36–47 and 48–67 in the 68-point scheme) and global landmark displacements, normalized by inter-ocular distance, with a threshold of $\geq$ 23:
\begin{equation}
    \text{change score} = 0.7 \times \text{inner diff} + 0.3 \times \text{overall diff},
    \label{change_score}
\end{equation}
where $\text{inner diff}$ is the mean absolute difference of the inner facial landmarks (indices 36–47 and 48–67), and $\text{overall diff}$ is the mean absolute difference across all 68 landmarks. For semantic diversity, we employ Qwen2.5-VL-32B to calculate a [0,1] normalized difference score between high-level visual features, retaining only pairs exceeding a threshold of 0.4.  Finally, CLIP~\citep{radford2021learning} is used for cross-frame identity verification, with a high similarity threshold of 0.9. For synthetic data, we estimate pitch and yaw from 68-point facial landmarks and retain only samples whose angular deviation from the instructed pose is within 10°.

\subsection{Expression instruction for real images}

\begin{figure}[htbp]
    \centering
    \includegraphics[width=0.85\linewidth]{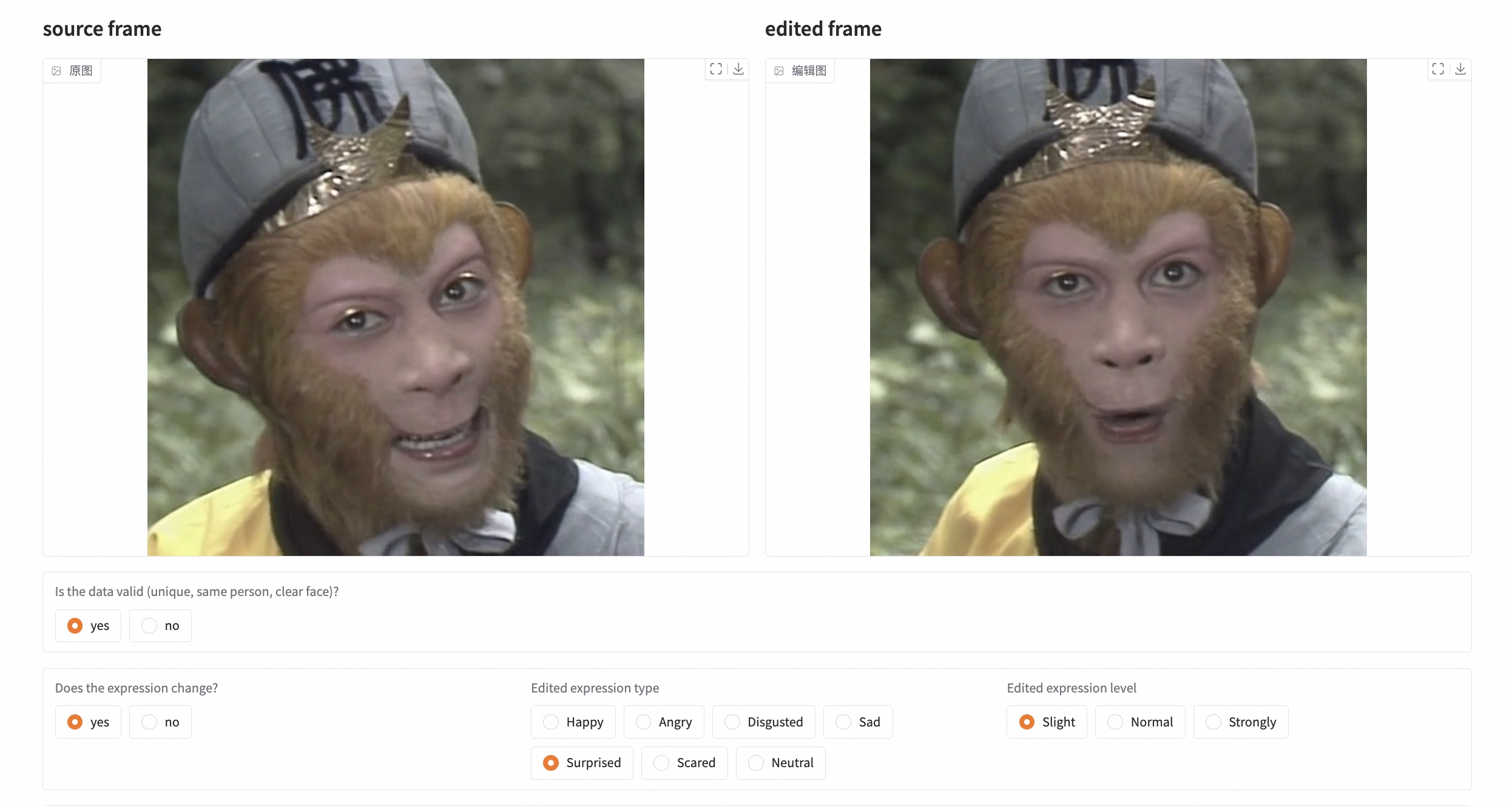}
    \caption{User interface for expression annotation.}
    \label{fig:annotation}
\end{figure}

\begin{figure}[!t]
    \centering
    \includegraphics[width=0.9\linewidth]{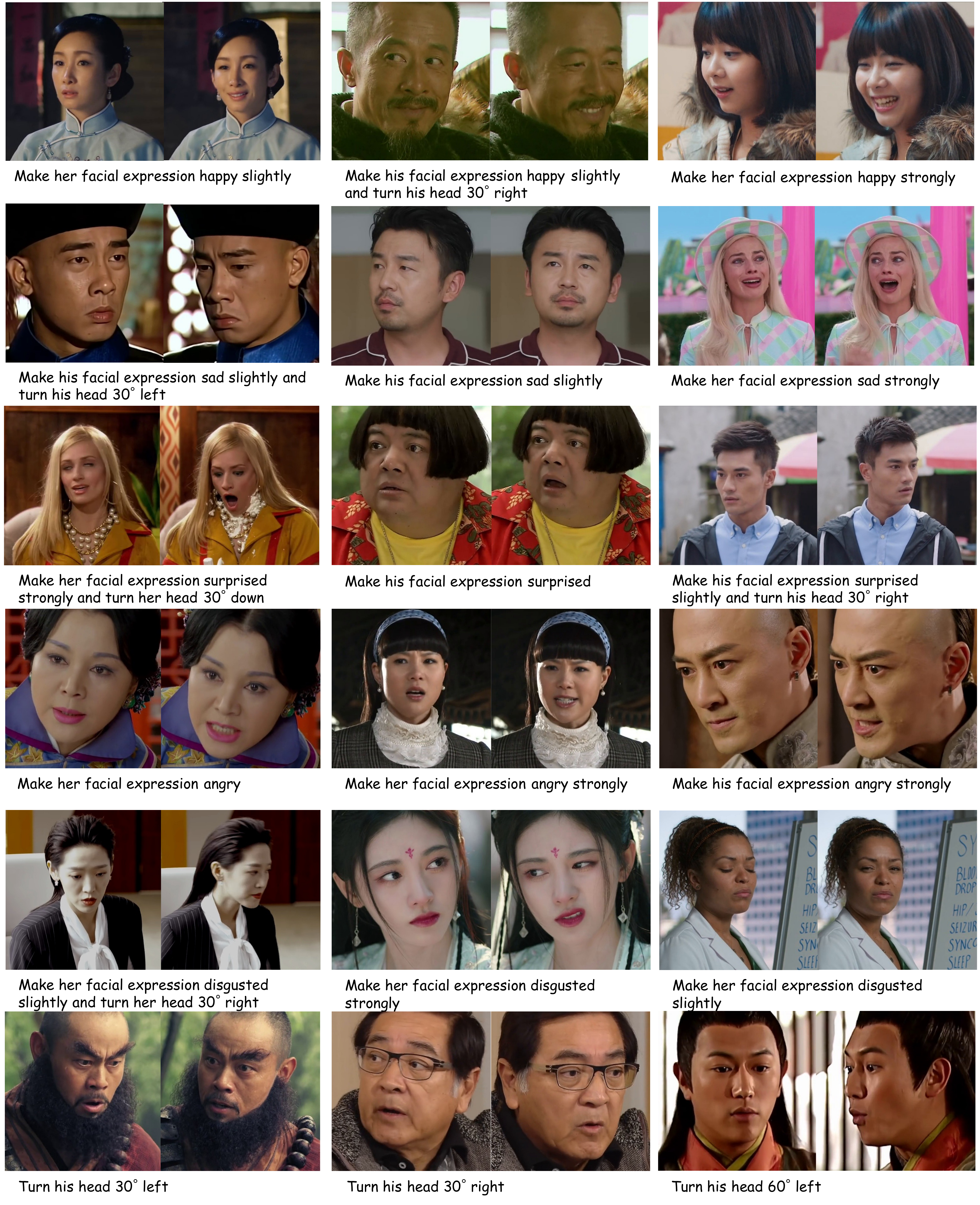}
    \caption{Examples from HFL-150K. Each triplet shows a source image, its instruction, and the corresponding target edit.}
    \label{fig:data_show}
\end{figure}

We observe that even Qwen2.5-VL-72B struggles to capture expression changes with sufficient granularity. To address this, we curate an 18K expression caption dataset with intensity labels via a crowdsourcing interface (Figure~\ref{fig:annotation}). Five annotators independently score each sample, and we retain only items with agreement from at least three raters. This corpus is used to fine-tune Qwen2.5-VL-7B, with 17K samples for training and 1K for evaluation. Evaluation adopts a hierarchical metric. Expression recognition receives 1 point only when the predicted category matches the ground truth. Intensity estimation receives 0.5 points only when the expression is correct and the predicted intensity matches the annotation. This cascading design penalizes intensity errors only after correct expression recognition. Table~\ref{exp_res} reports state-of-the-art results. The validated facial captioner is then used to generate expression instructions for real-world image pairs. Figure~\ref{fig:data_show} illustrates several training samples from HFL-150K.

\subsection{Diversity Statistics} Following standard practice in population studies~\citep{buolamwini2018gender}, we estimate age, gender, and skin tone for HFL-150K. The dataset contains 34.2$\%$ subjects aged 40 or above, 48.3$\%$ female and 51.7$\%$ male, and 23.1$\%$ dark‑skinned individuals, with the remaining subjects distributed across medium skin tones at 45.7$\%$ and light skin tones at 31.2$\%$. Intuitively, this indicates that the dataset is not overly concentrated on a narrow demographic, but instead offers a reasonably balanced coverage of different groups. We further observe that these proportions remain stable when we focus on challenging cases such as extreme head poses and strong facial expressions, suggesting that the demographic diversity is preserved even in hard scenarios.

\begin{table}[t]
\centering
\begin{minipage}[t]{0.48\linewidth}
\vspace{0pt}%
\centering
\setlength{\tabcolsep}{3.5pt}
\renewcommand{\arraystretch}{1.0}
\small
\captionof{table}{Comparison of captioning performance across expression granularity levels.}
\begin{tabular}{cc}
\toprule
Methods            & Accuracy \\ 
\midrule
Qwen2.5-VL-7B &   46.5\\
Qwen2.5-VL-72B &    56.9\\
\textbf{Ours}  &     67.3\\
\bottomrule
\end{tabular}
\label{exp_res}
\end{minipage}
\hfill
\begin{minipage}[t]{0.48\linewidth}
\vspace{0pt}%
\centering
\setlength{\tabcolsep}{3pt}
\renewcommand{\arraystretch}{1.0}
\small
\setlength{\tabcolsep}{2.5pt}
\renewcommand{\arraystretch}{1.0}
\small
\captionof{table}{Scaling analysis of HFL-150K.}
\begin{tabular}{ccccc}
\toprule
Training size           & SC~$\uparrow$ & VQ~$\uparrow$ & NA~$\uparrow$ & IP~$\uparrow$ \\ 
\midrule
20K &     0.784                   &    0.728            &     0.749 & 0.571              \\
50K &     0.813                  &    0.735            &     0.767 & 0.593              \\
100K &         0.828                     &       0.746               &   0.782             &  0.617                   \\
150K     &    0.832                  &  0.749                     &     0.805           &    0.634                 \\
\bottomrule
\end{tabular}
\label{tab:dataset}
\end{minipage}
\end{table}

\section{MORE EXPERIMENTS}
\subsection{Scaling Analysis of HFL-150K}
To assess HFL-150K’s scaling behavior, we train LaTo on stratified subsets of 20K, 50K, 100K, and 150K samples while preserving class distribution. All models are trained from scratch under identical settings. Table~\ref{tab:dataset} shows clear scaling trends: instruction adherence and visual quality improve rapidly at moderate scales (50K, 100K), whereas identity consistency and photorealism depend more strongly on larger, higher-quality data, with the full 150K set yielding the most robust gains. These results indicate that increasing data volume improves geometric modeling, which in turn strengthens identity preservation and natural appearance. Systematic expansion of the dataset via our pipeline is expected to further advance large-scale facial editing.

\subsection{User study} We conduct a user study on all 1,110 cases across the three benchmarks and evaluate Semantic Consistency, Visual Quality, and Identity Preservation using a 5‑point Likert scale for fine‑grained assessment. The detailed questionnaires are provided in Section ~\ref{D:3}. For a fair comparison, we evaluate Flux.1‑Kontext† and Step1X‑Edit†, both fine‑tuned on the HFL‑150K training set, and obtain 21 valid participant responses. As shown in Table~\ref{tab:user_study}, LaTo achieves an overall rating that is about 4.1$\%$ higher than Flux.1‑Kontext†, indicating that users consistently perceive LaTo’s edits as better aligned with the target semantics, visually more plausible, and more faithful to the original identity, and therefore more suitable for fine‑grained facial editing in practical applications.

\begin{table}[t]
\centering
\caption{User study results.}
\label{tab:user_study}
\begin{tabular}{lcccc}
\toprule
Methods & SC & VQ & IP & Overall \\
\midrule
Step1X-Edit†  & 0.728 & 0.810 & 0.746 & 0.761 \\
Flux.1-Kontext† & 0.712 & 0.824 & 0.774 & 0.770 \\
LaTo & 0.766 & 0.837 & 0.829 & 0.811 \\
\bottomrule
\end{tabular}
\end{table}

\subsection{Extension to Video-based Facial Animation} 
\begin{figure}[!t]
    \centering
    \includegraphics[width=0.9\linewidth]{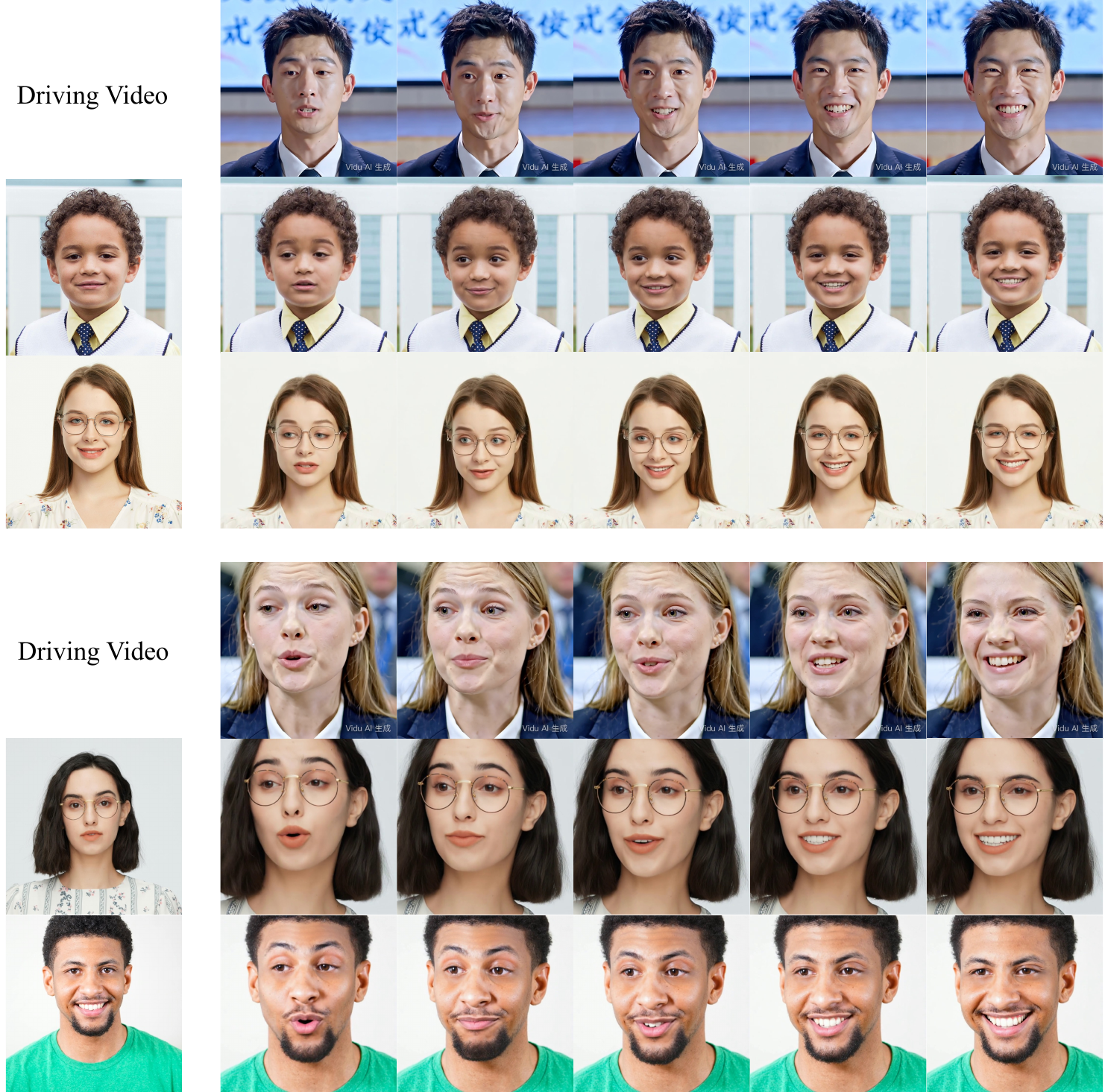}
    \caption{Examples of facial animation generated by LaTo. Given a cross‑identity driving video, LaTo can also achieve identity–geometry decoupling in the video domain, preserving the target identity while following the facial motions of the driving video.}
    \label{fig:video_show}
\end{figure}

We extend both the Landmark Tokenizer and the location‑mapping positional encoding to facial animation, enabling controllable video‑avatar generation.  The key differences from LaTo are as follows. For data, we construct a 16k high‑quality single‑portrait dataset from OpenHumanVid using similar filtering criteria, with each video annotated by landmark trajectories and descriptive captions. For the tokenizer, we generalize it to 3D with a temporal length of 81 frames. For fusion, we adopt Wan2.2‑TI2V‑5B~\citep{wan2025wan} as the base model and introduce an additional cross‑attention layer after the text‑to‑video attention. During training, we freeze all original weights and only train this new cross‑attention, which makes the adaptation efficient and avoids disrupting the pretrained text‑video alignment.  The visual results in Figure~\ref{fig:video_show} show that LaTo maintains robust identity preservation when editing with cross‑identity reference face videos, demonstrating strong scalability. However, the current system is reference‑video driven, and further work is needed to support interactive control similar to LaTo. These include developing a reliable landmark predictor for generating landmark sequences and identifying a more optimal fusion strategy, as the current additional cross‑attention may be suboptimal.

\subsection{Comparison of Landmark Predictors}
\begin{figure}[!t]
    \centering
    \includegraphics[width=0.95\linewidth]{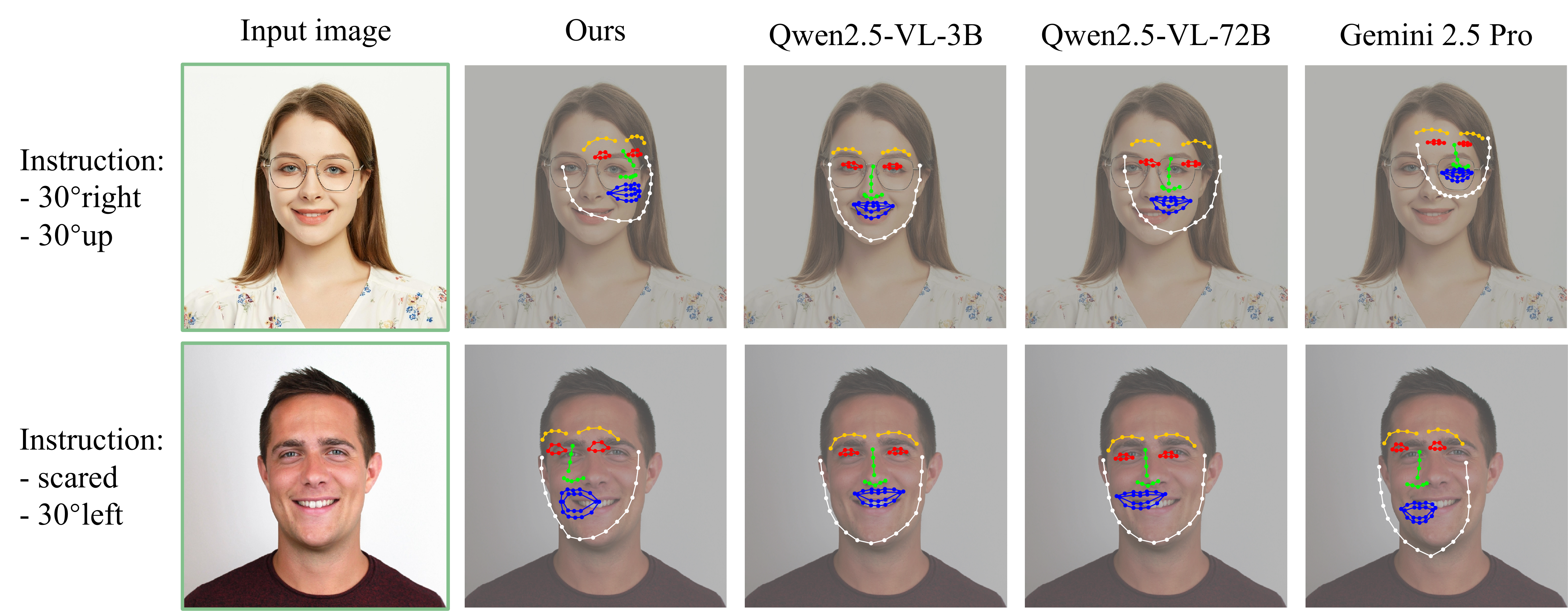}
    \caption{Qualitative comparison of our landmark predictor with existing VLMs.}
    \label{fig:compare_lp}
\end{figure}

Figure~\ref{fig:compare_lp} compares qualitative outputs across landmark predictors. Qwen2.5-VL produces unstable, low-magnitude updates. It under-responds to instructions that demand large pose or expression changes, yielding displacements clustered near the source landmarks, and it often ignores low-contrast regions, leaving jawline and cheek shifts static. Gemini 2.5 Pro better follows the requested direction but frequently distorts facial geometry. Common failure cases include global scaling that alters inter-ocular distance, jawline widening or compression, and asymmetric eyebrow motions not supported by the instruction. These behaviors suggest that general-purpose VLMs lack task-specific supervision for landmark reasoning. In contrast, our landmark predictor, fine-tuned for structured chain-of-thought landmark reasoning, accurately captures pose- and expression-induced landmark shifts while preserving facial shape, establishes a stable mapping between anatomical features and fine-grained linguistic instructions, and enables downstream edits that preserve identity while faithfully realizing the requested pose and expression changes.

\subsection{Design Choices for Unconditional Landmark Tokens}
\begin{figure}[!t]
    \centering
    \includegraphics[width=0.8\linewidth]{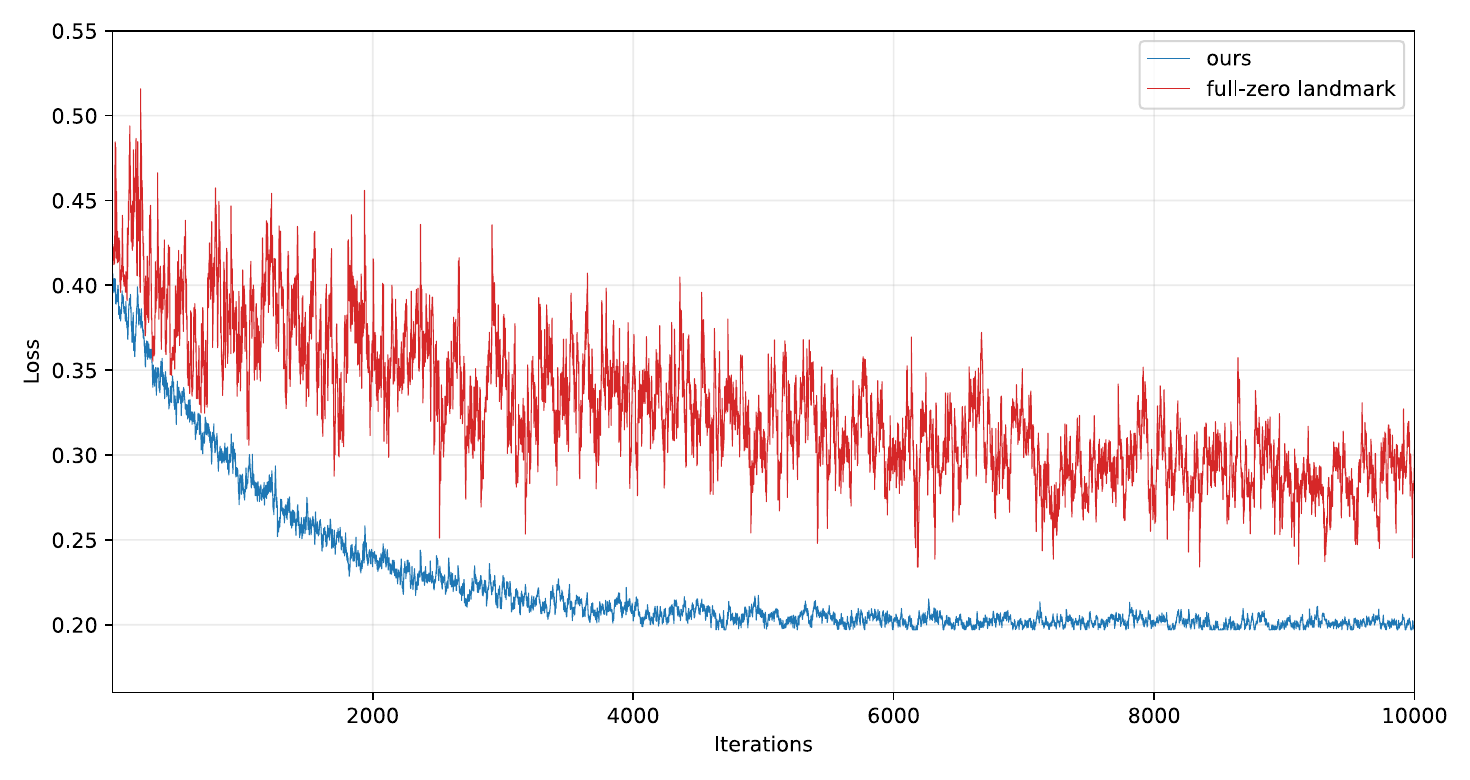}
    \caption{Training loss across iterations for different unconditional landmark token implementations.}
    \label{fig:loss}
\end{figure}

\begin{figure}[t]
\centering
\begin{minipage}[t]{0.42\textwidth}
\centering
\includegraphics[width=\linewidth]{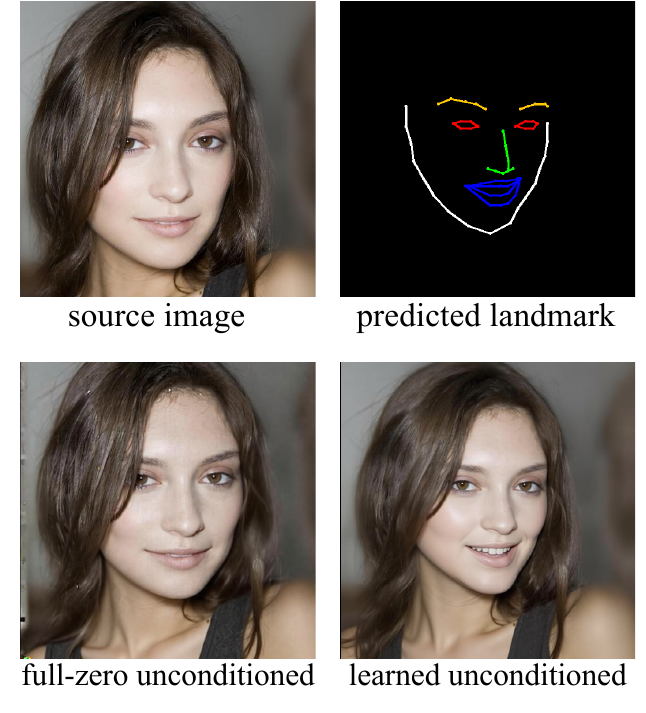}
\caption{Qualitative visualization of generated outputs using different unconditional landmark token strategies.}
\label{fig:cfg_diff}
\end{minipage}\hfill
\begin{minipage}[t]{0.42\textwidth}
\centering
\includegraphics[width=\linewidth]{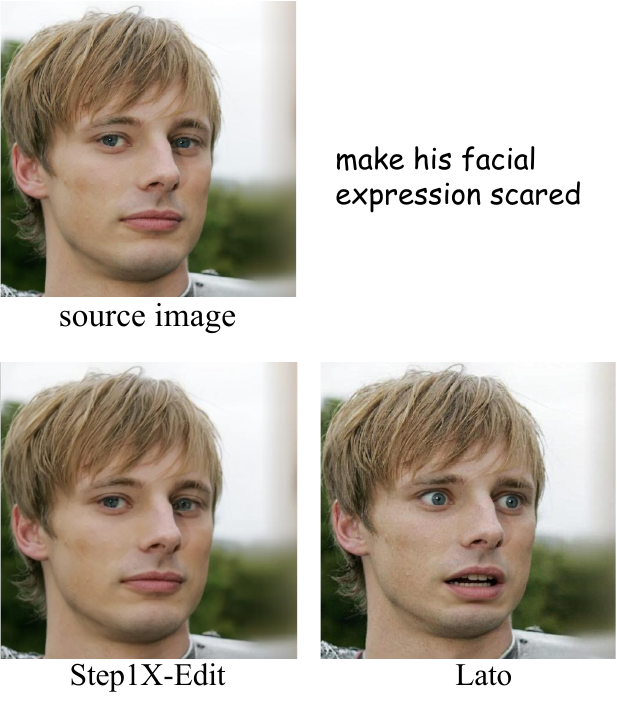}
\caption{Example of minor edits that contradict the editing instructions.}
\label{fig:rip}
\end{minipage}\hfill
\end{figure}

We investigate the trade-off between landmark alignment and instruction fidelity in the unconditional branch of classifier-free guidance. Two strategies are compared: a zero landmark list mapped to the tokenizer’s embedding space, and learnable unconditional landmark tokens. As shown in Figure~\ref{fig:loss}, the zero-embedding strategy exhibits unstable optimization with divergent losses and strong sensitivity to guidance scales. Zero vectors fail to represent an unconstrained landmark state, which conflicts with the dynamics of facial geometry. This mismatch leads to two characteristic failures: image-wide artifacts and weak alignment with the target landmarks and instruction (Figure~\ref{fig:cfg_diff}). In contrast, learnable unconditional tokens produce stable training, encode the unconstrained state explicitly, and integrate landmark guidance with instruction-based editing to achieve coherent cross-modal alignment.

\subsection{Rectified  Identity Preservation Score}
We find that identity evaluation can be biased when a model avoids making the requested edit. Even advanced metrics such as ArcFace or large VLMs report high similarity despite clear instruction violations. In Figure~\ref{fig:rip}, the instruction “make his facial expression scared” is ignored by Step1X-Edit. The output shows minimal change($\Delta SSIM = 0.05$) while ArcFace remains high at 0.98, which fails to reflect the editing intent. 
To address this limitation, we propose a rectified IP score that couples identity similarity with transformation magnitude and instruction compliance. As outlined in Algorithm~\ref{alg:1}, the Qwen2.5-VL first identifies facial regions in the source image and predicts the expected editing amplitude $ \varphi_{ins}$ from the instruction. We then measure the realized change $ \varphi_{real}$ using SSIM between the source and edited images and compute a discrepancy penalty $p$ with a factor $\alpha$ = 2. The rectified score $s_{rip}$ combines the identity score $s_{arc}$ with this penalty. For the example in Figure~\ref{fig:rip}: Step1X-Edit yields $\varphi_{ins} = 0.257, \varphi_{real} = 0.05, s_{arc}=0.984  \Rightarrow p =0.648, s_{rip} = 0.336$, while LaTo achieves $\varphi_{ins} = 0.257, \varphi_{real} = 0.341, s_{arc}=0.759 \Rightarrow p =0.065, s_{rip} = 0.694$. 

We validate the rectified IP score with a user study. We generate win–lose pairs for Step1X-Edit and LaTo using the original IP metric and the rectified IP metric. Participants judge whether the score difference aligns with their perception under the given instruction. The rectified IP score aligns with human preference with a 7\% error rate, which substantially outperforms ArcFace at 46\%.

\begin{algorithm}[!t]
\caption{Rectified Identity Preservation Score Calculation}
\label{alg:evaluation}
\begin{algorithmic}[1]
\REQUIRE Source image $ I_s $, edited image $ I_t $, editing instruction $ T $
\ENSURE Final score $ s_{rip} $

\STATE $ s_{arc} \gets \text{ArcFace}(I_s, I_t) $ \hfill $s_{arc} \in [0,1]$
\STATE $ \varphi_{ins} \gets \text{GetExpectedMagnitude}(I_s, T) $ \hfill $ \varphi_{ins} \in [0,1]$ \par
       \hfill $\text{via QwenVL prediction}$
\STATE $ \varphi_{real} \gets \text{SSIMDiff}(I_s, I_t) $ \hfill $\varphi_{real} \in [0,1]$
\STATE $ \alpha \gets 2 $, $ \epsilon \gets 1e-5 $ \hfill \text{Hyperparameters}
\STATE 
$$
p \gets (\frac{\varphi_{ins} - \varphi_{real}}{\varphi_{ins} + \epsilon}) ^\alpha
$$
\STATE $ s_{rip} \gets \max\left(0, s_{arc} -  p \right) $
\RETURN $s_{rip}$
\end{algorithmic}
\label{alg:1}
\end{algorithm}

\begin{figure}[!t]
    \centering
    \includegraphics[width=1\linewidth]{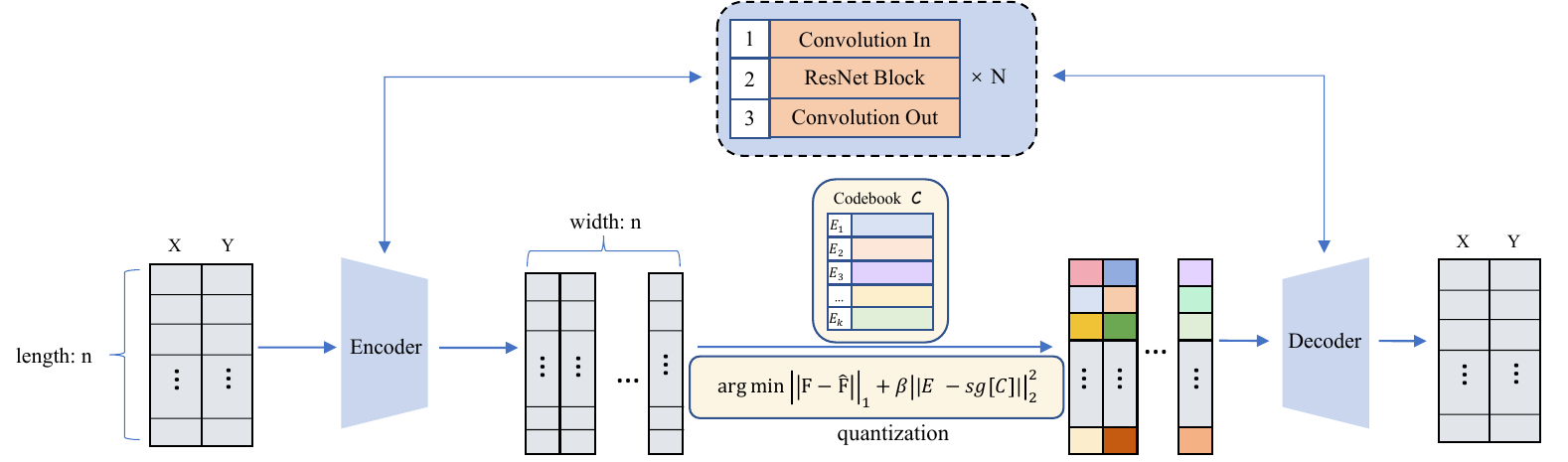}
    \caption{The architecture of the landmark tokenizer.}
    \label{fig:lm_tok}
\end{figure}

\subsection{Extended Qualitative Comparisons}
\begin{figure}[!t]
    \centering
    \includegraphics[width=1\linewidth]{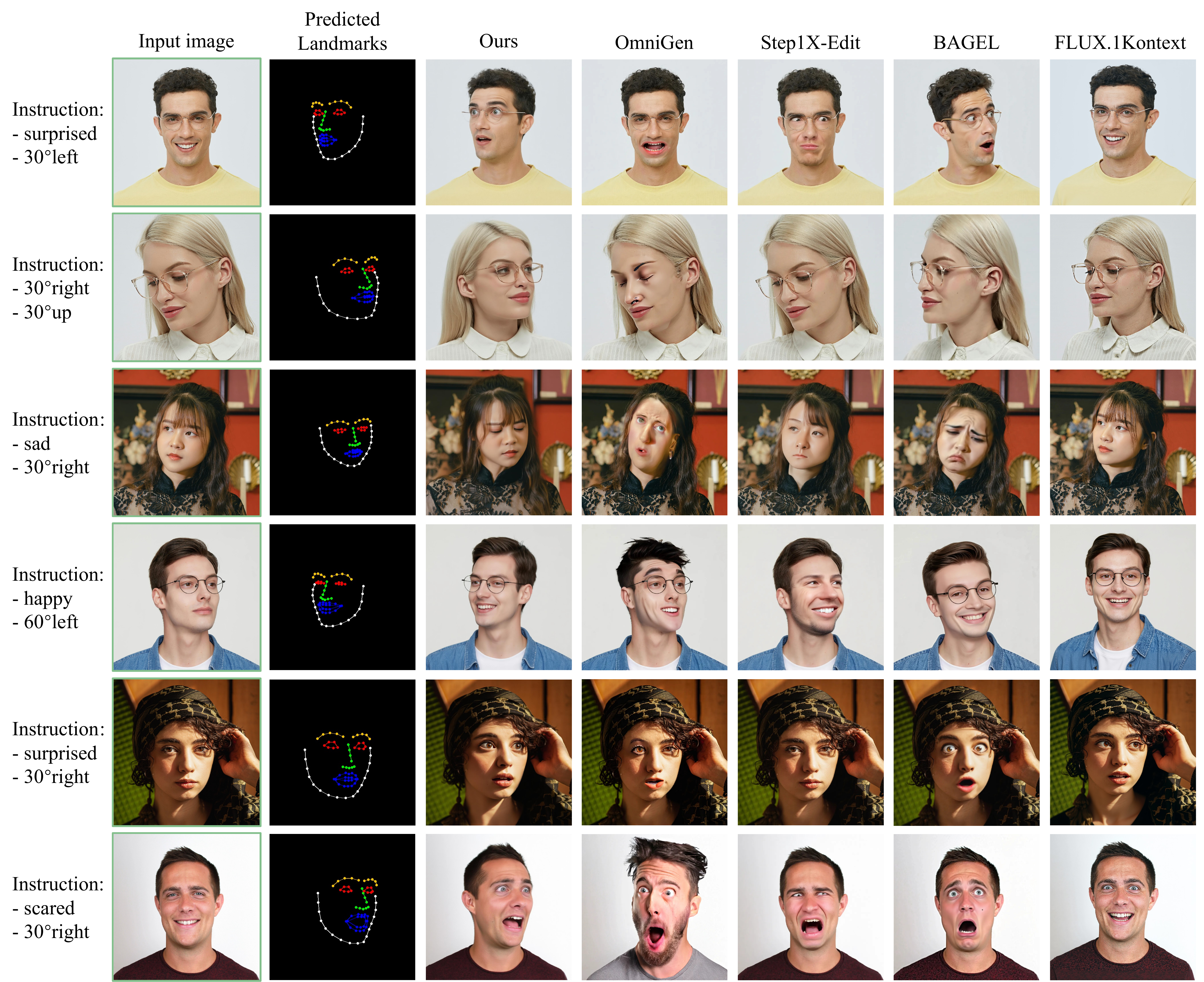}
    \caption{Additional comparisons with state-of-the-art editing methods.}
    \label{fig:comparison}
\end{figure}

Figure~\ref{fig:comparison} presents additional qualitative results under diverse instructions. LaTo follows the specified edits while preserving identity and facial geometry. Baselines often produce artifacts, unnatural appearance, or landmark drift. These results further demonstrate the effectiveness of our design for controllable face editing.

\section{Additional Methodological Details}
\subsection{landmark tokenizer}
Figure~\ref{fig:lm_tok} shows the detailed architecture of the landmark tokenizer. To quantify codebook diversity, we computed cosine similarities for all pairs of code vectors and report aggregate statistics: mean $\approx$ 0.007 and standard deviation $\approx$ 0.019 on the test set. The near-zero mean and low dispersion indicate that the code vectors are close to orthogonal, confirming that the tokenizer induces a discrete embedding space with strong vector distinctiveness and minimal redundancy.

We also evaluated codebook sizes of 1,024, 8,192, and 16,384, as well as a no‑quantization ablation. Table~\ref{tab:codebook_size} reports landmark reconstruction performance on HFL‑150K. For PSNR evaluation, we render the landmarks into images. Removing quantization forces the tokenizer to operate on continuous and inconsistent facial features, which degrades performance. This confirms that discrete and general landmark tokens provide more stable geometric representations and improve generalization. Although larger codebooks offer higher nominal capacity, the 16,384‑entry codebook resulted in unstable optimization and significant codebook underutilization, which reduced its effective capacity. The 8,192‑entry codebook achieved the best overall performance. We anticipate further gains through improved codebook utilization. 

\begin{table}[t]
\centering
\caption{Ablation of codebook size.}
\label{tab:codebook_size}
\begin{tabular}{lccc}
\toprule
Codebook Size & Utilization (\%) & L1 Distance & PSNR \\
\midrule
no-quantization & - & 7.04 & 19.11 \\
1024 & 81.42 & 5.32 & 20.47 \\
8192 & 70.39 & 3.67 & 21.33 \\
16384 & 44.75 & 4.51 & 20.61 \\
\bottomrule
\end{tabular}
\end{table}

\subsection{Structured Chain-of-Reasoning for the Landmark Predictor}
\ref{lst:cot_prompt} presents the prompt used to elicit a structured chain of reasoning conditioned on the source image, the instruction, and the target landmark. The template enforces a machine-parsable schema that includes:

\begin{tcolorbox}[
    colback=white,
    colframe=blue!20!black,
    title=\textbf{Facial Landmark Prediction Chain-of-Thought Prompt},
    fonttitle=\bfseries,
    enhanced,
    breakable,
    sharp corners,
    boxrule=1pt,
    left=6pt,
    right=6pt,
    top=6pt,
    bottom=6pt,
    label=lst:cot_prompt
]
\noindent
\texttt{\#\#\# Role and Task} \\
You are an expert in facial geometry, computer vision, and human facial expression analysis. Your task is to generate a detailed, explicit, step-by-step Chain-of-Thought (CoT) reasoning process that shows exactly how to derive the edited facial landmark coordinates (target output) by reasoning from the following four inputs:

\begin{enumerate}
    \item The input image of a human face.
    \item The original facial landmarks of the face (given as precise coordinate arrays, typically grouped by facial regions such as JAW/BROWS, NOSE, EYES, MOUTH).
    \item The natural language editing instruction describing how the face should be transformed (e.g., change in expression, head pose, or other features).
    \item The edited facial landmarks (target result coordinates after applying the transformation).
\end{enumerate}

Your output is NOT to be a direct mapping or a simple coordinate difference. Instead, it must be a structured internal reasoning narrative that reflects the plausible human-expert thought process in getting from the original state (using 1, 2, 3) to the edited state (4). This reasoning should:

\begin{itemize}
    \item Begin with an \textbf{Initial State Analysis} using the input image and original landmarks to describe the current facial pose, expression, and alignment.
    \item Include anatomical and geometric interpretation of the original landmarks and how they correspond to key facial features.
    \item \textbf{Decompose the editing instruction} into precise, separate transformations (e.g., rigid 3D head rotation, non-rigid facial muscle changes).
    \item Analyze \textbf{which landmark groups} are affected by each transformation and how (both direction and magnitude of expected movement in image coordinate space).
    \item Consider both \textbf{global rigid transformations} (e.g., head rotation, translation, scaling effects due to perspective) and \textbf{local non-rigid deformations} (e.g., eye widening, eyebrow lifting, mouth shape change).
    \item Provide quantitative or semi-quantitative estimates of coordinate shifts (in X/Y pixels) that are consistent with the original and final coordinates.
    \item Explain how perspective and foreshortening influence the new positions after a head turn.
    \item Show how expression changes (like smiling $\to$ scared) affect local geometry, including muscle tension, lip curvature, jaw drop, eyelid movement, eyebrow displacement.
    \item Maintain \textbf{identity coherence} in the resulting shape: distances and proportions in rigid facial structures should remain consistent with physical reality.
    \item Explicitly connect initial landmark positions (from 2) to final ones (from 4) through the reasoning steps.
\end{itemize}

The final Chain-of-Thought should be \textbf{comprehensive, logically progressive, and reflect expert-level spatial reasoning}---not just descriptions of differences. Use a numbered step-by-step format (e.g., Step 1, Step 2...) and include both qualitative anatomical analysis and approximate numerical transformation where applicable. Always explain \textit{why} each change happens given the transformation described in the instruction, and ensure the reasoning is consistent with realistic facial kinematics. Output the CoT reasoning directly, without any irrelevant descriptions.

\vspace{1em}
\noindent
\texttt{\#\#\# Inputs for your reasoning will be presented in the following format:}
\begin{enumerate}
    \item \{Input image\}
    \item \{Original Landmarks JSON\}
    \item \{Editing Instruction\}
    \item \{Edited Landmarks JSON\}
\end{enumerate}

\vspace{1em}
\noindent
\texttt{\#\#\# Chain-of-Thought Reasoning you should produce:} \\
Step 1: Initial State Analysis.
\begin{itemize}
    \item Analyze the Visual Reference Image to understand the face's starting condition.
    \item Current Pose:
    \item Current Expression:
    \item Correlate the visual features with the provided Initial Normalized Landmark Coordinates to build a mental model of the face.
\end{itemize}

Step 2: Instruction Decomposition and Kinematic Analysis.
\begin{itemize}
    \item Break down the instruction \textless Your Input Editing Instruction\textgreater{} into primary anatomical movements.
    \item Primary Action(s):
    \item Key Facial Parts Affected:
    \item Kinematic Chain Reasoning: (Describe how these parts move together as a 2D projection within the 512x512 normalized coordinate space.)
\end{itemize}

Step 3: Quantitative Transformation Estimation (in the 512x512 space).
\begin{itemize}
    \item Translate the qualitative reasoning into quantitative coordinate shifts for landmark groups. All estimations refer to the 512x512 normalized canvas.
    \item Example Transformation for 'Turn right and smile':
    \begin{itemize}
        \item \texttt{jawline (points 0-16)}: "Left points (0-8) will shift right (positive X) by about 25--40 units. Right points (9-16) will also shift right, but by a smaller amount, maybe 10--20 units."
        \item \texttt{nose (points 27-35)}: "All points will shift right (positive X). Point 33 (tip) will move the most, maybe 30 units. Point 27 (root) will move the least, maybe 8 units."
        \item \texttt{mouth (points 48-67)}: "Points 48 and 54 will move up (negative Y) by ~15 units and horizontally apart by ~10 units each. The center of the upper lip (point 51) will rise (negative Y) by ~8 units."
    \end{itemize}
    \item Identity Constraint: "All transformations must be cohesive. The relative distances within rigid groups (like the nose bridge) should be largely preserved, while deformable areas (like the mouth) change shape according to the expression. The overall transformation should look like the same person."
\end{itemize}

\vspace{1em}
\noindent
\texttt{\#\#\# Input:} \\
1. [Input image] \\
2. Original Landmarks JSON:
\begin{lstlisting}[style=jsonstyle]
{"JAW/BROWS": [[160, 198], [160, 226], [164, 247], [171, 267], [178, 291], [191, 312], [202, 322], [219, 336], [250, 346], [277, 339], [298, 326], [315, 315], [329, 295], [336, 271], [339, 247], [346, 226], [346, 198], [174, 174], [184, 167], [198, 167], [212, 167], [226, 174], [274, 174], [284, 171], [298, 167], [315, 171], [325, 178]], "NOSE": [[250, 202], [246, 222], [246, 236], [246, 250], [233, 257], [236, 260], [246, 264], [257, 260], [267, 257]], "EYES": [[191, 198], [202, 195], [212, 195], [226, 202], [215, 205], [202, 202], [274, 202], [284, 198], [298, 198], [305, 202], [298, 205], [284, 205]], "MOUTH": [[212, 281], [222, 278], [239, 278], [250, 278], [257, 278], [274, 278], [288, 284], [274, 298], [260, 305], [246, 308], [236, 305], [222, 298], [215, 281], [236, 284], [250, 284], [260, 284], [288, 284], [260, 291], [246, 295], [236, 291]]}
\end{lstlisting}
3. Editing Instruction: turn his/her head 30 degrees to the right and 30 degrees up \\
4. Edited Landmarks JSON:
\begin{lstlisting}[style=jsonstyle]
{"JAW/BROWS": [[192, 198], [197, 220], [202, 242], [211, 263], [226, 280], [248, 291], [275, 298], [301, 303], [324, 303], [341, 298], [350, 285], [357, 267], [362, 249], [364, 231], [365, 212], [361, 195], [355, 179], [233, 164], [243, 151], [260, 143], [277, 142], [294, 147], [317, 146], [327, 139], [337, 138], [346, 141], [351, 151]], "NOSE": [[310, 167], [316, 176], [323, 185], [330, 195], [305, 217], [315, 217], [324, 218], [329, 216], [334, 214]], "EYES": [[255, 179], [265, 170], [275, 169], [283, 176], [275, 179], [265, 180], [320, 174], [329, 166], [338, 165], [344, 173], [338, 175], [329, 175]], "MOUTH": [[281, 249], [298, 238], [315, 232], [325, 234], [332, 231], [340, 234], [342, 245], [341, 255], [335, 262], [326, 264], [316, 264], [299, 260], [286, 249], [315, 242], [325, 241], [332, 241], [339, 245], [333, 250], [326, 252], [316, 253]]}
\end{lstlisting}

\vspace{1em}
\noindent
\texttt{\#\#\# Output:}
\end{tcolorbox}

\subsection{Face Editing Questionnaire}
\label{D:3}

Thank you for completing the following rating of image editing results. Please spend about 15 seconds on each image, consider the source image and the editing instruction, and rate the result based on the following dimensions:

\textbf{1. Semantic Consistency} \\
Does it match the text prompt (e.g., ``Smiling slightly and turn her head left 30 degrees'')? \\
1: No match at all. \\
2: Very weak or incorrect interpretation (e.g., wrong expression or pose direction). \\
3: Partial match — correct type but inaccurate intensity or rotation degree. \\
4: Mostly accurate — small deviation in expression strength or head angle. \\
5: Perfectly matches the prompt in both expression and head pose.

\medskip

\textbf{2. Visual Quality} \\
How photorealistic and artifact-free is the generated face? \\
1: Severely distorted or unrealistic — obvious artifacts, broken facial structure, or non-human appearance. \\
2: Low quality — blurry, noisy, or with noticeable distortions (e.g., warped eyes, asymmetric features). \\
3: Acceptable but flawed — overall plausible, but has mild blurriness, texture glitches, or minor asymmetry. \\
4: High quality — sharp and realistic, with only subtle imperfections hard to notice at a glance. \\
5: Excellent quality — indistinguishable from a real photo; no visible artifacts or distortions.

\medskip

\textbf{3. Identity Preservation } \\
Is it the same person as the source? \\
1: Completely different person. \\
2: Unlikely the same — key facial features (e.g., eye spacing, nose shape, jawline) don’t match. \\
3: Uncertain — could be the same or different. \\
4: Likely the same — minor structural differences, but core identity preserved. \\
5: Definitely the same person — identity is unmistakable.

\section{Limitations}
Though LaTo is effective for fine-grained editing and identity preservation, the complex architecture of the advanced base model limits its ability to support efficient on-the-fly processing, so we rely on future advances in accelerating such base models by the community. Moreover, we observe that for input images with extreme head poses caused by heavy occlusion, landmark estimation becomes less reliable, leading to slight degradation in identity preservation. We plan to incorporate 3D-aware priors and richer occlusion annotations to mitigate these limitations.

\end{document}